\documentclass{article}

\PassOptionsToPackage{numbers,sort,compress}{natbib}
\usepackage{open_iclr,times}

\usepackage{tikz} % draw timeline
\usepackage{adjustbox}
\usepackage[utf8]{inputenc} % allow utf-8 input
\usepackage[T1]{fontenc}    % use 8-bit T1 fonts
\usepackage{url}            % simple URL typesetting
\usepackage{booktabs}       % professional-quality tables
\usepackage{microtype}      % microtypography        
\usepackage{algorithm}
\usepackage{algpseudocode}
\usepackage{listings}
\usepackage{overpic}
\usepackage{wrapfig}
\usepackage{graphicx}
\usepackage{colortbl}
\usepackage{makecell}
\usepackage{multirow}
\usepackage{tabularx}
\usepackage{verbatim}
\usepackage{nicefrac}
\usepackage{empheq}         % For the boxed equations
\usepackage{adjustbox}

\usepackage[pagebackref=true,breaklinks=true,colorlinks,bookmarks=false,citecolor=blue,linkcolor=blue]{hyperref}

\usepackage{amsthm}
\usepackage{amssymb}
\usepackage{amsmath}
\usepackage{amsfonts}       % blackboard math symbols
\usepackage{nicefrac}       % compact symbols for 1/2.
\usepackage{mathtools}

\usepackage[capitalize]{cleveref}
\crefname{equation}{Eq.}{Eq.}
\crefname{figure}{Fig.}{Fig.}
\crefname{table}{Tab.}{Tab.~}
\crefname{section}{Sec.}{Sec.~}
\crefname{algorithm}{Alg.}{Alg.~}
\crefname{thm}{Theorem}{Theorem~}
\crefname{lemma}{Lemma}{Lemma~}
\crefname{appendix}{Appendix}{Appendix~}

\theoremstyle{plain}

\theoremstyle{definition}

\theoremstyle{remark}

\usepackage{xcolor}
\definecolor{tomato}{rgb}{1.0, 0.39, 0.28}
\definecolor{cornflowerblue}{rgb}{0.39, 0.58, 0.93}

\def\ie{\textit{i.e.,~}}

\def\sota{state-of-the-art~}

\usepackage{amsmath,amsfonts,bm}

\def\eqref#1{equation~\ref{#1}}
\def\1{\bm{1}}

\def\rvx{{\mathbf{x}}}
\def\rvy{{\mathbf{y}}}

\def\mI{{\bm{I}}}

\DeclareMathAlphabet{\mathsfit}{\encodingdefault}{\sfdefault}{m}{sl}
\SetMathAlphabet{\mathsfit}{bold}{\encodingdefault}{\sfdefault}{bx}{n}

\def\gD{{\mathcal{D}}}

\def\gN{{\mathcal{N}}}

\def\gU{{\mathcal{U}}}

\newcommand{\E}{\mathbb{E}}

\newcommand{\R}{\mathbb{R}}

\DeclareMathOperator*{\argmin}{arg\,min}

\def\f{f_\theta}

\def\x{\rvx}
\def\y{\rvy}

\def\d{{\mathrm{d}}}
\def\dist{{\mathbf{d}}}

\def\imgnet{ImageNet 64$\times$64}
\def\dino{$\text{FD}_{\text{DINOv2}}$}

\title{Consistency Models Made Easy}

\author{%
    Zhengyang Geng$^1$
    \And
    Ashwini Pokle$^1$
    \And
    William Luo$^2$
    \And
    Justin Lin$^1$
    \And
    J. Zico Kolter$^1$
    \And
    \vspace{-0.4cm} \\
    $^1$CMU \quad $^2$Westlake University
}

\begin{document}

\maketitle

\begin{abstract}

Consistency models (CMs) offer faster sampling than traditional diffusion models, but their training is resource-intensive. For example, as of 2024, training a state-of-the-art CM on CIFAR-10 takes one week on 8 GPUs. In this work, we propose an effective scheme for training CMs that largely improves the efficiency of building such models. Specifically, by expressing CM trajectories via a particular differential equation, we argue that diffusion models can be viewed as a special case of CMs. We can thus fine-tune a consistency model starting from a pretrained diffusion model and progressively approximate the full consistency condition to stronger degrees over the training process. Our resulting method, which we term Easy Consistency Tuning (ECT), achieves vastly reduced training times while improving upon the quality of previous methods: for example, ECT achieves a 2-step FID of 2.73 on CIFAR10 within 1 hour on a single A100 GPU, matching Consistency Distillation trained for hundreds of GPU hours. Owing to this computational efficiency, we investigate the scaling laws of CMs under ECT, showing that they obey the classic power law scaling, hinting at their ability to improve efficiency and performance at larger scales. Our \href{https://github.com/locuslab/ect}{code} is publicly available, making CMs more accessible to the broader community.

\end{abstract}
\section{Introduction}
Diffusion Models (DMs)~\citep{ho2020denoising,ddim}, or Score-based Generative Models (SGMs)~\citep{song2020score,scoresde}, have vastly changed the landscape of visual content generation with applications in images~\citep{rombach2021high,saharia2022image,ho2022cascaded,dhariwal2021diffusion,hatamizadeh2023diffit,ramesh2021zero}, videos~\citep{videoworldsimulators2024,blattmann2023stable,bar2024lumiere,ho2022imagen,gupta2023photorealistic}, and 3D objects~\citep{poole2022dreamfusion,wang2024prolificdreamer,lee2024dreamflow,chen2024vp3d,babu2023hyperfields}. 
DMs progressively transform a data distribution to a known prior distribution (e.g. Gaussian noise) according to a stochastic differential equation (SDE)~\citep{scoresde} and train a model to denoise noisy observations. 
Samples can be generated via a reverse-time SDE that starts from noise and uses the trained model to progressively denoise it.
However, sampling from a DM naively requires hundreds to thousands of model evaluations due to the curvature of the diffusion sampling trajectory~\citep{Karras2022edm}, making the entire generative process slow. 
Many approaches have been proposed to address this issue,  
including training-based techniques such as  distillation~\citep{Luhman2021KnowledgeDI,salimans2022progressive,di,gu2023boot,sauer2023adversarial,geng2024one,yin2023one,nguyen2023swiftbrush}, adaptive compute architectures for the backbone model~\citep{moon2023early,tang2023deediff}, as well as training-free methods such as fast samplers~\citep{kong2021fast,lu2022dpm,zhang2022fast,zhou2023fast,xue2024sa} or interleaving small and large backbone models during sampling~\citep{pan2024t}.
However, the speedup achieved by these sampling techniques usually comes at the expense of the quality of generated samples.

Consistency Models (CMs)~\citep{cm} are a new family of generative models, closely related to diffusion models, that have demonstrated promising results as faster generative models. 
These models learn a mapping between noise and data, and all the points of the sampling trajectory map to the same initial data point.
Owing to this condition, consistency models are capable of generating high-quality samples in 1-2 model evaluations.
The best such models so far, built using improved Consistency Training (iCT)~\citep{ict}, have pushed the quality of images generated by 1-step CMs trained from scratch to a level comparable with SoTA DMs using thousands of steps for sampling.   
Unfortunately, CMs remain time-consuming and practically challenging to train: the best practice takes many times longer than similar-quality DMs while involving complex hyperparameter choices in the training process.  
In total, this has substantially limited the uptake of CMs within the community.

In this work, we introduce a differential perspective on consistency models, leading to the formulation of the differential consistency condition in continuous time. This insight reveals the link between diffusion models and consistency models, viewing DMs as a special case of CMs with loose discretization.
This observation motivates us to smoothly interpolate from DM to CM by progressively tightening the consistency condition, bootstrapping pretrained DMs to 1-step CMs w/o using extra frozen teachers.
We term this strategy as \emph{Easy Consistency Tuning (ECT)}, which includes diffusion pretraining as a special stage of the continuous time training schedule.

ECT significantly improves both training efficiency and performance. On \imgnet~\citep{imagenet}, ECT achieves superior 1-step and 2-step sample quality compared to the prior art. Similarly, on CIFAR-10~\citep{cifar} 2-step sample quality of ECT surpasses previous methods.  The total cost of the pretraining and tuning scheme requires only $\nicefrac{1}{4} \sim \nicefrac{1}{3}$ of the computational resources (FLOPs) used by the current \sota method, iCT~\citep{ict}, while the tuning stage can be remarkably lightweight, typically accounting for $10\%$ or less of the overall cost and further benefiting from scaling.

Leveraging ECT's computational efficiency, we conduct the first study into the scaling behaviors of CMs, revealing the classic power law scaling for model size, FLOPs, and training compute. 
The scaling also suggests a sweet spot of using smaller few-step CMs over larger 1-step CMs in certain scenarios. 
This computational efficiency enables us to explore the design space of CMs by using the tuning stage as a proxy. Notably, tuning findings, such as weighting functions, can improve the pretraining stage and, in turn, enhance the overall pretraining + tuning pipeline for CMs.

In short, we summarize our contributions as follows:
\begin{itemize}
    \item  We develop Easy Consistency Tuning (ECT), a pretraining + tuning scheme for training CMs in continuous time, demonstrating significant efficiency and performance gains compared to the current best practices for training CMs.
    \item We investigate CMs' scaling behaviors for the first time and reveal the classic power law.
    \item We explore the design space of CMs through ECT, introducing the continuous-time schedule and better weighting functions for CMs.
\end{itemize}

\section{Preliminaries}\label{sec:preliminaries}

\paragraph{Diffusion Models.} Let $p_{\text{data}}(\rvx_0)$ denote the data distribution. Diffusion models (DMs) perturb this distribution by adding monotonically increasing i.i.d. Gaussian noise with standard deviation $\sigma(t)$ from $t=0$ to $T$ such that $p_t(\rvx_t | \rvx_0) = \gN(\rvx_0, \sigma^2(t)\mI)$, and $\sigma(t)$ is chosen such that $\sigma(0) = \sigma_{\text{min}}$ and $\sigma(T) = \sigma_{\text{max}}$.
This process is described by the following SDE~\citep{scoresde}
\begin{equation}
    \d \rvx_t = \mathbf{f}(\rvx_t, t) \d t + g(t) \d \mathbf{w}_t \label{eqn:ito-sde},
\end{equation}
where $\mathbf{w}$ is the standard Wiener process, $\mathbf{f}(\cdot, t): \R^d \rightarrow \R^d$ is the drift coefficient, and $g(\cdot): \R \rightarrow \R$  is the diffusion coefficient. 
Samples can be generated by solving the reverse-time SDE starting from $t=T$ to $0$ and sampling $\rvx_T \sim \gN(0, \sigma^2_\text{max}\mI)$.
\citep{scoresde} show that this SDE has a corresponding ODE, called the probability flow ODE (PF-ODE), whose trajectories share the same marginal probability densities as the SDE. We follow the notation in \citep{Karras2022edm} to describe the ODE as
\begin{equation}
    \d\rvx_t = -\dot{\sigma}(t) \sigma(t) \nabla_{\rvx_t} \log p_t(\rvx_t)\d t,
\end{equation}
where $\nabla_{\rvx_t} \log p_t(\rvx_t)$ denotes the score function. Prior works~\citep{Karras2022edm,cm} set $\sigma(t) = t$ which yields
\begin{equation}
    \dfrac{\d \rvx_t}{\d t} =  -t \nabla_{\rvx_t} \log p_t(\rvx_t) = \dfrac{(\rvx_t - f(\rvx_t,t))}{t},\label{eq:karras-ode}
\end{equation}
where $f(\rvx_t,t)$ denotes a denoising function that predicts clean image $\rvx_0$ given noisy image $\rvx_t$. We will follow this parametrization in the rest of this paper. Note that time is same as noise level with this parametrization, and we will use these two terms interchangeably.

\paragraph{Consistency Models.} CMs are built upon the PF-ODE in \cref{eq:karras-ode}, which establishes a bijective mapping between data distribution and noise distribution. 
CMs learn a \textit{consistency function} $f(\x_t,t)$ that maps the noisy image $\x_t$ back to the clean image $\x_0$
\begin{equation}
f(\x_t,t) = \x_0.
\end{equation}
Note that the consistency function needs to satisfy the boundary condition at $t = 0$. Prior works \citep{cm,ict,Karras2022edm} impose this boundary condition by parametrizing the CM as
\begin{equation}
f_\theta(\x_t, t) = c_{\text{skip}}(t) \, \x_t + c_{\text{out}}(t) \, F_\theta (\x_t, t),
\end{equation}
where $\theta$ is the model parameter, $F_\theta$ is the network to train, and $c_{\text{skip}}(t)$ and $c_{\text{out}}(t)$ are time-dependent scaling factors such that $c_{\text{skip}}(0) = 1$, $c_{\text{out}}(0) = 0$. 
This parameterization guarantees the boundary condition by design. 
We discuss specific choices of $c_{\text{skip}}(t)$ and $c_{\text{out}}(t)$ in \cref{sec:experiment-details}. 

During training, CMs first discretize the PF-ODE into $N-1$ subintervals with boundaries given by  $t_{\text{min}} = t_1 < t_2 < \hdots < t_N = T$. 
The model is trained on the following CM loss, which minimizes a metric between adjacent points on the sampling trajectory
\begin{equation}
    \argmin_{\theta} \E\left[w(t_i) d(f_\theta(\rvx_{t_{i+1}}, t_{i+1}), f_{\theta^-}(\tilde{\rvx}_{t_{i}}, t_{i}))\right].
\end{equation}
Here, $d(\cdot, \cdot)$ is a metric function, the $f_\theta$ indicates a trainable neural network that is used to learn the consistency function, $f_{\theta^-}$ indicates an exponential moving average (EMA) of the past values of $f_\theta$, and $\tilde{\rvx}_{t_{i}} = \rvx_{t_{i+1}} - (t_{i} - t_{i+1}) t_{i+1} \nabla_{\rvx_{t_{i+1}}} \log p_{t_{i+1}}(\rvx_{t_{i+1}})$. 
Further, the discretization curriculum $N$ should be adaptive and tuned during training to achieve good performance.

In the seminal work, \citep{cm} use Learned Perceptual Similarity Score (LPIPS)~\citep{lpips} as a metric function, set $w(t_i) = 1$ for all  $t_i$, and sample $t_i$ according to the sampling scheduler by ~\citep{Karras2022edm}:
$t_i = \left(t_{\text{max}}^{1/\rho} + \frac{i}{N-1}(t_{\text{min}}^{1/\rho} - t_{\text{max}}^{1/\rho})\right)^\rho$ for $i \in \gU[1, N-1]$ and $\rho=0.7$. Further, the score function 
$\nabla_{\x_{t}} \log p(\x_t)$
 can either be estimated from a pretrained diffusion model,
which results in Consistency Distillation (CD), 
or can be estimated with an unbiased score estimator 
\begin{equation}
\label{score-est}
\nabla_{\x_{t}} \log p(\x_{t})
= \E\left[\nabla_{\x_{t}} \log p(\rvx_{t} | \rvx_{0}) \bigg\vert \rvx_t\right]
= \E\left[- \frac{\rvx_t - \rvx_0}{t^2} \bigg\vert \rvx_t\right],
\end{equation}
which results in consistency training (CT).

The follow-up work, iCT~\citep{ict}, introduces several improvements that significantly improve training efficiency as well as the performance of CMs. 
First, the LPIPS metric, which introduces undesirable bias in generative modeling, is replaced with a Pseudo-Huber metric. 
Second, the network  $f_{\theta^-}$ does not maintain an EMA of the past values of $f_\theta$. Third, iCT replaces the uniform weighting scheme $w(t_i) = 1$ with $w(t_i) = \frac{1}{t_{i+1} - t_i}$. 
Further, the scaling factors of noise embeddings and dropout are carefully selected. 
Fourth, iCT introduces a complex discretization curriculum during training:
\begin{equation}
\label{schedule}
\begin{aligned}
& N(m) = \min(s_0 2^{\left\lfloor \frac{m}{M'} \right\rfloor}, s_1) + 1, \quad M' = \left\lfloor \frac{ M }{\log_2 \left\lfloor \frac{s_1}{s_0} \right\rfloor + 1} \right\rfloor,
\end{aligned}
\end{equation}
where $m$ is the current number of iterations, $M$ is the total number of iterations, $\sigma_{\max}$ and $\sigma_{\min}$ is the largest and smallest noise level for training, $s_0 = 10$ and $s_1 = 1280$ are hyperparameters.
Finally, during training, iCT samples $i \sim p(i) \propto
\text{erf} \left( \frac{\log(t_{i+1}) - P_{\text{mean}}}{\sqrt{2} P_{\text{std}}} \right) -
\text{erf} \left( \frac{\log(t_i) - P_{\text{mean}}}{\sqrt{2} P_{\text{std}}} \right)$ from a discrete Lognormal distribution, where $P_{\text{mean}} = -1.1$ and $P_{\text{std}}=2.0$. 

\section{Probing consistency models}
We will first introduce the differential consistency condition and the loss objective based on this condition. 
Next, we will analyze this loss objective and highlight the challenges of training CMs with it.
Based on this analysis, we present our method, Easy Consistency Tuning (ECT).
ECT is a simple, principled approach to efficiently train CMs to meet the (differential) consistency condition. 
The resulting CMs can generate high-quality samples in 1 or 2 sampling steps.

\subsection{Differential Consistency Condition}

As stated in \cref{sec:preliminaries}, CMs learn a \textit{consistency function} $f(\x_t,t)$ that maps the noisy image $\x_t$ back to the clean image $\x_0$:  $f(\x_t,t)=\x_0$.
Instead, by taking the time derivative of both sides, given by the differential form
\begin{equation}
    \frac{\mathrm{d}f}{\mathrm{d}t} = \frac{\mathrm{d}}{\mathrm{d}t} \x_0 = 0. \label{eq:differential-form}
\end{equation}

However, the differential form $\frac{\mathrm{d}f}{\mathrm{d}t}=0$ alone is not sufficient to guarantee that the model output will match the clean image, 
as there exist trivial solutions where the model maps all the inputs to a constant value, such as 
$f(\x_t, t) \equiv 0$. 
To eliminate these collapsed solutions, \citep{cm,ict} impose the boundary condition for $f(\x_t, t) = \x_0$ via model parameterization:

\begin{empheq}[box=\fbox]{align}
f(\x_t, t) = \x_0 
\Leftrightarrow 
\frac{\mathrm{d}f}{\mathrm{d}t} = 0, f(\x_0, 0) = \x_0. \label{eq:boundary-cond}
\end{empheq}

This boundary condition $f(\x_0, 0) = \x_0$ ensures that the model output matches the clean image when the noise level is zero. Together, the differential form in \cref{eq:differential-form} and the boundary condition define the \textit{differential consistency condition} in \cref{eq:boundary-cond}, or \textbf{\textit{consistency condition}} in short.

\paragraph{Finite Difference Approximation.}
To learn the consistency condition, we discretize the differential form $\frac{\mathrm{d}f}{\mathrm{d}t} = 0$ using a finite-difference approximation:
\begin{equation}
0 = \frac{\mathrm{d}f}{\mathrm{d}t} \approx \frac{f_\theta(\x_t) - f_\theta(\x_r)}{t - r} 
\end{equation}
where $\mathrm{d}t \approx \Delta t = t - r$, $t > r >= 0$, and $f_\theta(\x_t)$ denotes $f_\theta(\x_t, t)$. 
For a given clean image $\x_0$, we produce two perturbed images $\x_t$ and $\x_r$ using the same perturbation direction $\boldsymbol{\epsilon} \sim p(\boldsymbol{\epsilon})$ at two noise levels $t$ and $r$.
Specifically,
we compute $t$ and $r$ according to the forward process of CMs i.e., $\x_t = \x_0 + t \cdot \boldsymbol{\epsilon}$ and $\x_r = \x_0 + r \cdot \boldsymbol{\epsilon}$.

To satisfy the consistency condition, we minimize the distance between $f_\theta(\x_t)$ and $f_\theta(\x_r)$. For the boundary condition, we optimize $f_\theta(\x_t)$ to align with the clean image $\x_0$. For noise levels $t > r > 0$, we freeze $f_{\operatorname{sg}(\theta)}(\x_r)$ using the stop-gradient operator $\operatorname{sg}$ and optimize $f_\theta(\x_t)$ on higher variance inputs to align with self teacher's prediction $f_{\operatorname{sg}(\theta)}(\x_r)$ on lower variance inputs~\citep{chen2021exploring}.
\paragraph{Loss Function.} Given the discretization, the training objective for CMs can be formulated as:
\begin{equation}\label{eq:CMcost}
\underset{\theta}{\arg \min} \; \mathbb{E}_{\x_0,\boldsymbol{\epsilon},t}\left[w(t, r)\dist(f_\theta(\x_t), f_{\operatorname{sg}(\theta)}(\x_r))
\right],
\end{equation}
where we can extract a weighting function $w(t, r) = \frac{1}{t-r}$, $\x_t = \x_0 + t \cdot \boldsymbol{\epsilon}$, $\x_r = \x_0 + r \cdot \boldsymbol{\epsilon}$ using a \textit{shared noise direction} $\boldsymbol{\epsilon} \sim p(\boldsymbol{\epsilon})$, and $\dist(\cdot, \cdot)$ is a metric function.
%
% In practice, we set $\dist(\cdot, \cdot)$ to the squared $L_2$ metric. Further, we discuss an interesting relation between the Pseudo-Huber metric proposed in \citep{ict} and the weighting function in ~\cref{sec:ect}.
This formulation generalizes \citet{cm}'s discrete training schedule to continuous time.
To regularize the model, dropout masks should also be consistent across $f_\theta(\x_t)$ and $f_\theta(\x_r)$ by setting a shared random seed, ensuring that the noise levels $t$ and $r$ are the only varying factors.

\subsection{The "Curse of Consistency" and its Implications}
\label{sec:curse-of-consistency}
The objective in \cref{eq:CMcost} can be challenging to optimize when $\Delta t \to 0$. This is because the prediction errors from each discretization interval accumulate, leading to slow training convergence or, in the worst case, divergence. 
To further elaborate, consider a large noise level $T$. We first follow the notation of CT~\citep{cm} and split the noise horizon $[0,T]$ into $N$ smaller consecutive subintervals. We can error bound the error of $1$-step prediction as
\begin{equation}\label{eq:consistency-error}
\| f_\theta(\x_T) - \x_0\| 
\leq 
\sum_i^N\| f_\theta(\x_{t_i}) - f_\theta(\x_{r_i}) \|
\leq 
N e_{\max},
\end{equation}
\begin{wrapfigure}{r}{0.4\textwidth}
    \centering
    \vspace{-5pt}
    \includegraphics[width=0.4\textwidth]{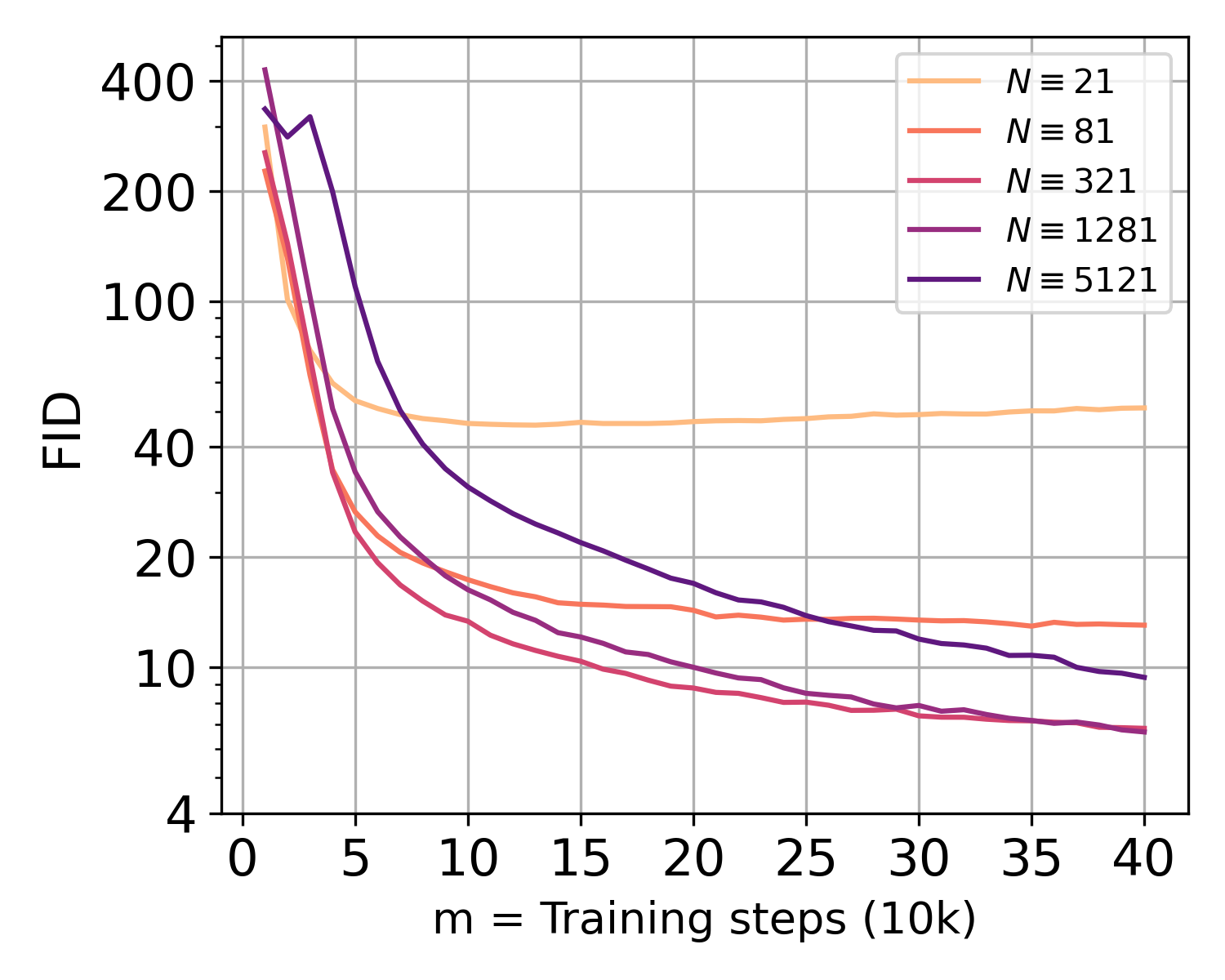}
    \vspace{-10pt}
    \caption{The "Curse of Consistency": The consistency condition holds at $\Delta t = \mathrm{d}t$. However, the training dynamics converges more slowly and is less stable as $\Delta t \to \d t$ (\ie $N \to \infty$).}
    \label{fig:curse}
    \vspace{-3pt}
\end{wrapfigure}
where $r_1 = 0 < t_1 = r_2 < \cdots < t_{N-2} = r_N < t_N = T $, and $e_{\max} = \max_i \| f_\theta(\x_{t_i}) - f_\theta(\x_{r_i}) \|$, for $i = 1, \cdots, N$. 
Ideally, we want both $N$ and $e_{\max}$ to be small so that this upper bound is small. But in practice, there is a trade-off between these two terms.
As $\Delta t_i = (t_i - r_i) \to 0$, $e_{\max}$ decreases because $\x_{t_i}$ and $\x_{r_i}$ will be close, and it is easier to predict both $f_\theta(\x_{r_i})$ and  $f_\theta(\x_{t_i})$. 
However, $N$ will increase as $\Delta t_i \to 0$. 
In contrast, for a large $\Delta t_i$, vice-versa holds true. 
It is difficult to theoretically estimate the rate at which $e_{\max}$ decreases when $\Delta t_i = t_i - r_i \to 0$, as it depends on the optimization process---specifically, how effectively we train the model to minimize the consistency error in each interval. 
If $e_{\max}$ decreases more slowly than $N$, the product of the two can increase instead, resulting in a worse prediction error $\| f_\theta(\x_T) - \x_0\|$.

The insight from the above observation is that when training from scratch by strictly following the differential form $\nicefrac{\mathrm{d}f}{\mathrm{d}t} = 0$ with a tiny $\Delta t_i \approx 0$, the resulting model might converge slowly due to accumulated consistency errors from each interval $\Delta t_i$. 
To investigate this hypothesis, we conducted experiments training a series of CMs using different fixed numbers of intervals $N$, and corresponding $\Delta t$ values for the consistency condition, as illustrated in \cref{fig:curse}. 
Our observations align with \citet{ict}, with the key distinction that we use a fixed $N$ for this analysis, as this approach provides better insight on how much the precise approximation of the differential consistency condition matters and isolates the effect of discretization errors.

\subsection{Easy Consistency Tuning (ECT)}
\label{sec:ect}
The discussion in \cref{sec:curse-of-consistency} highlighted the training instability issue that can arise when we naively train CMs from scratch and directly optimize for the differential consistency condition with the loss objective in \cref{eq:CMcost} while directly following $\Delta t \approx 0$.
In this section, we propose several strategies that largely alleviate the aforementioned training issues and improve the efficiency of CMs. We term this approach Easy Consistency Tuning (ECT), as it effectively balances training stability and model performance while simplifying the CM training process. ECT follows a two-stage approach: diffusion pretraining, followed by consistency tuning, which we will detail in the following subsections.

\paragraph{Diffusion Pretraining + Consistency Tuning.} Drawing inspiration from iCT's adaptive discrete-time schedule~\citep{ict}, we start ECT with a large $\Delta t$, and gradually shrink $\Delta t \to 0$. In our problem setup, Since $t > r \geq 0$, we have the largest possible $\Delta t = t$ with $r=0$, which yields
\begin{equation}
\underset{\theta}{\arg \min} \; \| f_\theta(\x_t) - f_{\operatorname{sg}(\theta)}(\x_r) \| 
= \| f_\theta(\x_t) - f_{\operatorname{sg}(\theta)}(\x_0) \| 
= \| f_\theta(\x_t) -\x_0 \|.
\end{equation}
Training a model with this loss is \emph{identical} to diffusion model/Score SDE~\citep{ho2020denoising,scoresde}. This observation suggests a learning scheme that smoothly interpolates from DMs $\Delta t = t$ to CMs $\Delta t = \mathrm{d}t$ by gradually shrinking $\Delta t \to \mathrm{d}t$ during training by gradually tightening the consistency condition.
With this reasoning, diffusion pretraining can be considered as a special case of consistency training with a loose discretization of the consistency condition. 
Therefore, in practice, we start ECT with a pretrained diffusion model, resulting in a training scheme of pretraining+tuning.
Another benefit of this initialization is that during training, especially in the initial stages, it ensures good targets $f_{\operatorname{sg}(\theta)}(\x_r)$ in the loss objective, avoiding trivial solutions.
%
% Update later
% \asp{We note that iCT begins training with a small $N$, such as $N$=$10$. 
% Morever, extrapolating to $N$=$1$ leads to predicting the solution of the MMSE estimator, which results in suboptimal performance. 
% This is precisely where the advantage of the training strategies proposed in ECT will become evident, as ECT smoothly interpolates from DMs to CMs. 
% Please see the results in [Add FID DINO table] in the appendix. - TODO: Why does starting from N=1 not harm ECT but harms iCT? Give a short justification too.}

We highlight two advantages of this learning scheme: 
1. The pretraining+tuning scheme of ECT outperforms iCT's training-from-scratch approach with lower overall computational cost (see \cref{sec:comparison-of-training-schemes}).
2. Tuning serves as an efficient proxy for exploring the CM design space. Given a pretrained diffusion model, insights gained during tuning can be applied to improve the pretraining stage, resulting in a more refined overall CM training pipeline.
Please refer to \cref{sec:additional-experimental-results} for more details.

\paragraph{Continuous-time Training Schedule.}

\begin{algorithm}
\caption{Easy Consistency Tuning (ECT)}
\label{alg:ect}
\begin{algorithmic}[t]
\State \textbf{Input:} Dataset $\gD$, a pretrained diffusion model $\phi$, mapping function $p(r \mid t, \mathrm{Iters})$, weighting function $w(t)$.
\State \textbf{Init:} $\theta \leftarrow \theta_{\phi}$, $\mathrm{Iters} = 0$.
\Repeat
    \State $\mathrm{Sample}$ $\x_0 \sim \mathcal{D}$, $\boldsymbol{\epsilon} \sim p(\boldsymbol{\epsilon})$, $t \sim p(t)$, $r \sim p(r \mid t, \mathrm{Iters})$
    \State $\mathrm{Compute}$ $\x_t = \x_0 + t \cdot \boldsymbol{\epsilon}$, $\x_r = \x_0 + r \cdot \boldsymbol{\epsilon}$, $\Delta t = t-r$
    \State $L(\theta) = w(t) \cdot \dist(f_\theta(\x_t), f_{\operatorname{sg}(\theta)}(\x_r))$ \Comment{$\mathrm{sg}$ is stop-gradient operator}
    \State $\theta \leftarrow \theta - \eta \nabla_\theta L(\theta)$
    \State $\mathrm{Iters} = \mathrm{Iters} + 1$
\Until{$\Delta t \to \mathrm{d}t$}
\Return $\theta$ \Comment{ECM}
\end{algorithmic}
\end{algorithm}
\vspace{-5pt}
We investigate the design principles of a continuous-time schedule whose "boundary" condition yields standard diffusion pretraining, \ie constructing training pairs of $r=0$ for all $t$ at the beginning. Note that this is unlike the discrete-time schedule used in iCT (See \cref{sec:preliminaries} and \cref{sec:design-choices}). 
We consider overlapped intervals for consistency models, which allows for factoring $p(t, r) = p(t) \, p(r|t)$ and continuous sampling of infinite $t$ from noise distribution $p(t)$, for instance, $\mathrm{LogNormal}(P_{\text{mean}}, P_{\text{std}})$, and $r \sim p(r | t)$. 

We refer to $p(r | t)$ as the mapping function. 
Since we need to shrink $\Delta t \to \d t$ as the training progresses, we augment the mapping function to depend on training iterations, $p(r|t, \text{iters})$, to control $\Delta t = (t - r) \to \d t$.
We parametrize the mapping function $p(r|t, \text{iters})$ as
\begin{equation}  
\frac{r}{t}  
= 1 - \frac{1}{q^a} n(t)
= 1 - \frac{1}{q^{\lfloor\mathrm{iters}/d\rfloor}} n(t),
\label{eq:mapping-fn}
\end{equation}
where we take $n(t) = 1 + k \,\sigma(-bt) = 1 + \frac{k}{1 + e^{bt}}$ with $\sigma(\cdot)$ as the sigmoid function, $\mathrm{iters}$ refers to training iterations. In general, we set $q > 1$, $k=8$ and $b=1$. 
Since $r\geq0$, we also clamp $r$ to satisfy this constraint.
At the beginning of training, this mapping function produces $\nicefrac{r}{t} = 0$, which recovers the diffusion pretraining.
We discuss design choices of this function in \cref{sec:design-choices}.

\vspace{-3pt}
\paragraph{Choice of metric.} iCT uses the Pseudo-Huber metric~\citep{huber} to mitigate the perceptual bias caused by the LPIPS metric~\citep{lpips}. 
When taking a careful look at this metric, we reveal that this metric offers adaptive per-sample scaling of the gradients, which reduces the variance of gradients during training.
Specifically, the differential of the Pseudo-Huber loss can be decomposed into two terms: an adaptive scaling factor $w(\Delta)$ and the differential of the squared $L_2$ loss. We discuss this in more detail in \cref{sec:design-choices}. 
Thus, we disentangle the weighting function from the loss metric as it provides greater flexibility. In our experiments, we retain the squared $L_2$ metric used in DMs and explore alternate choices of adaptive weighting terms that reduce the variance of gradients during training. Please refer to \cref{sec:additional-experimental-results} for further discussions.

\vspace{-3pt}
\paragraph{Weighting function.} 
Weighting functions usually lead to a substantial difference in performance in DMs, and the same holds true for CMs. 
Substituting the mapping function $p(r|t)$ in \cref{eq:mapping-fn} into the weighting function in \cref{eq:CMcost}, we get $w(t, r) = \nicefrac{1}{(t - r)} = q^a \cdot \nicefrac{1}{tn(t)}$, where $q^a$ is the scaling factor to maintain the loss scales when $\Delta t \to 0$, and $\nicefrac{1}{tn(t)}$ is the weighting akin to DMs. 
This couples the weighting function with the mapping function $p(r|t)$. Instead, we consider decoupled weighting functions without relying on $p(r|t)$.
Motivated by the adaptive scaling factor that appears in Pseudo-Huber loss (See \cref{sec:design-choices} for more details), we rewrite the weighting function as 
\begin{equation}
w(t) = \Bar{w}(t) \cdot w(\Delta) = \Bar{w}(t) \cdot \frac{1}{(\|\Delta \|_2^2 + c^2)^{\nicefrac{1}{2}}},
\end{equation}
where $\Delta = f(\rvx_t) - f(\rvx_r)$. We define $\Bar{w}(t)$ as timestep weighting, while $w(\Delta)$ as adaptive weighting.
The adaptive weighting improves training efficiency with the $L_2$ metric because, as $\Delta \to 0$ (usually happens when $t \to 0$), this weighting $ w(\Delta)$ upscales gradients to avoid vanishing gradient during learning of fine-grained features, while $c$ mitigates potential numerical issues when $\Delta \approx 0$.
We direct the reader to \cref{tab:wt-imgnet} in \cref{sec:additional-experimental-results} for a detailed overview of various choices of $\Bar{w}(t)$ and $w(\Delta)$ considered in this work.
In general, we notice that CMs' generative capability greatly benefits from weighting functions that control the variance of the gradients across different noise levels. 
We provide a brief summary of the core steps of ECT in \cref{alg:ect}.

\vspace{-5pt}
\section{Experiments} \label{sec:experiments}

This section compares different learning schemes and investigates scaling laws of ECT, while more experiments on design choices and scaling are shown in \cref{sec:additional-experimental-results}. We evaluate the efficiency and scalability of ECT on two datasets: CIFAR-10~\citep{cifar} and ImageNet $64 \times 64$~\citep{imagenet}.
We measure the sample quality using Fréchet Inception Distance (FID)~\citep{fid} and Fréchet Distance under the DINOv2 model~\citep{oquab2023dinov2} (\dino)~\citep{fd_dino} and sampling efficiency using the number of function evaluations (NFEs). We also indicate the relative training costs of each of these methods. Implementation details can be found in \cref{sec:experiment-details}.

\subsection{Comparison of Training Schemes}
\label{sec:comparison-of-training-schemes}
\begin{figure}[t!]
    \centering
    \includegraphics[width=\linewidth]{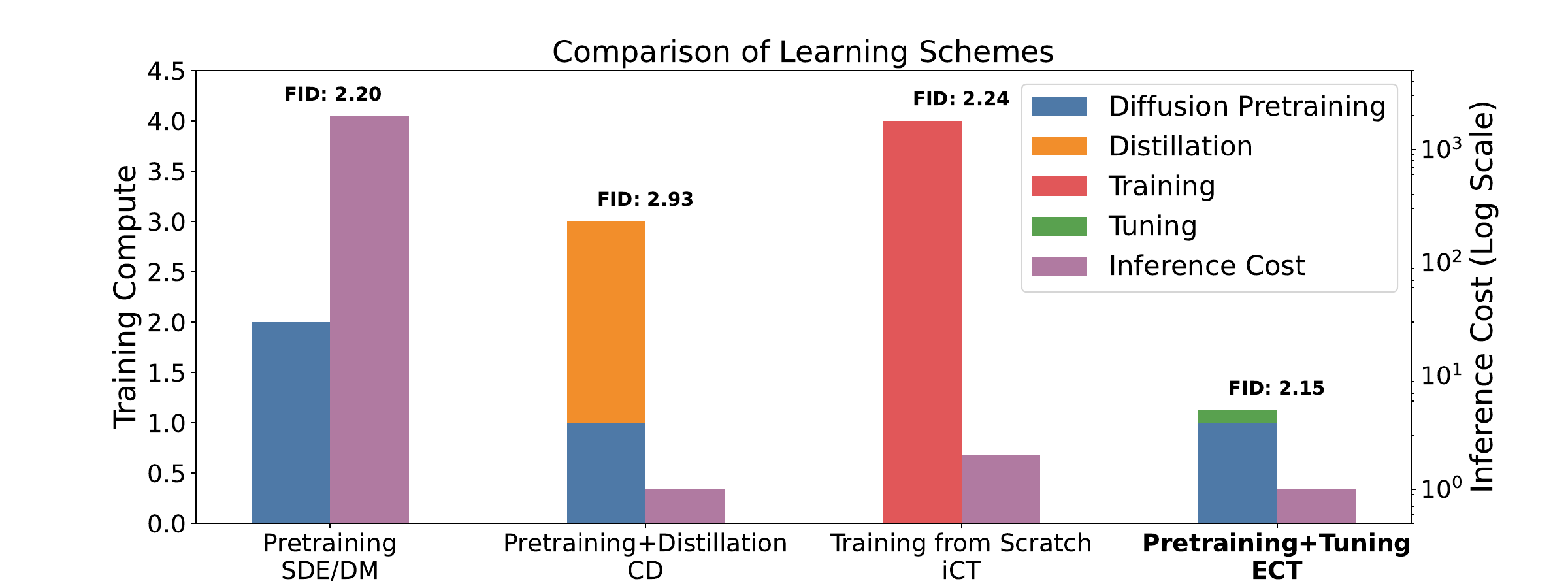}
    \caption{Comparison of training schemes for the diffusion-consistency family on CIFAR-10. Without relying on distillation from frozen diffusion teachers or extra adversarial supervision, ECT surpasses Consistency Distillation (CD)~\citep{cm} and Consistency Models trained from scratch (iCT)~\citep{ict} using $\nicefrac{1}{4}$ of the total training cost. ECT also significantly reduces the inference cost to $\nicefrac{1}{1000}$ compared to Diffusion Pretraining (Score SDE/DMs) while maintaining comparable sample quality.}
    \label{fig:efficiency}
    \vspace{-10pt}
\end{figure}

We compare CMs trained with ECT (denoted as ECM) against state-of-the-art diffusion models, advanced samplers for diffusion models, distillation methods such as consistency distillation (CD)~\citep{cm}, and improved Consistency Training (iCT)~\citep{ict}. 
The key results are summarized in \cref{fig:efficiency}. We show the training FLOPs, inference cost, and generative performance of the four training schemes. 

\begin{figure}[t!]
    \vspace{10pt}
    \centering
    \includegraphics[scale=0.68]{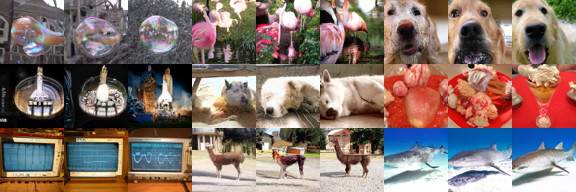}
    \caption{Scaling up training compute and model sizes results in improved sample quality on \imgnet. Each triplet (left-to-right) has $2$-step samples from ECM-S trained with $12.8\mathrm{M}$ images, ECM-S trained with $102.4\mathrm{M}$ images, and ECM-XL trained with $102.4\mathrm{M}$ images.}
    \label{fig:imgnet-scaling-samples}
    \vspace{-10pt}
\end{figure}

\paragraph{Score SDE/Diffusion Models.}
We compare ECMs against Score SDE~\citep{scoresde}, EDM ~\citep{Karras2022edm}, and EDM with DPM-Solver-v3~\citep{zheng2024dpm}. 
$2$-step ECM, which has been fine-tuned for only $100$k iterations, matches Score SDE-deep, with $2\times$ model depth and $2000$ NFEs, in terms of FID. 
As noted in \cref{fig:efficiency}, ECM only requires $\nicefrac{1}{1000}$ of its inference cost and latency to achieve the same sample quality.
$2$-step ECM fine-tuned for $100$k iterations outperforms EDM with advanced DPM-Solver-v3 (NFE=$10$). 

\paragraph{Diffusion Distillation.}
We compare ECT against Consistency Distillation (CD) \citep{cm}, a SoTA approach that distills a pretrained DM into a CM. As shown in \cref{table:ect-cifar10-imagenet64}, ECM significantly outperforms CD on both CIFAR-10 and \imgnet. We note that ECT is free from the errors of teacher DM and does not incur any additional cost of running teacher DM. 
$2$-step ECM outperforms $2$-step CD (with LPIPS~\citep{lpips}) in terms of FID ($2.20$ vs $2.93$) on CIFAR-10 while using around $\nicefrac{1}{3}$ of training compute of CD.
\paragraph{Consistency training from scratch.} Improved Consistency Training (iCT)~\citep{ict} is the SoTA recipe for training a consistency model from scratch without inferring the diffusion teacher. 
Compared to training from scratch, ECT rivals iCT-deep using $\nicefrac{1}{4}$ of the overall training compute ($\nicefrac{1}{8}$ in the tuning stage) and $\nicefrac{1}{2}$ of the model size as shown in \cref{fig:efficiency} and \cref{table:ect-cifar10-imagenet64}.

\begin{table}[t!]
% \vspace{10pt}
\caption{Generative performance on unconditional CIFAR-10 and class-conditional \imgnet. We use a budget of $12.8\mathrm{M}$ training images (batch size 128 and 100k iterations) for ECMs. $^\star$ stands for a budget of $102.4\mathrm{M}$ training images (batch size 1024 and 100k iterations) on \imgnet.}
\vspace{5pt}
\centering
\small
\begin{minipage}[t]{0.49\linewidth}
\centering
\setlength{\tabcolsep}{5pt} 
\renewcommand{\arraystretch}{1.00} % 
\begin{adjustbox}{max width=\textwidth}
    \begin{tabular}{@{}lccc@{}}
    \toprule
    \multicolumn{3}{c}{CIFAR-10} \\ 
    \midrule
    Method & FID$\downarrow$ & NFE$\downarrow$ \\ 
    \midrule
    {\color{gray} Diffusion Models} & ~ & ~ \\ 
    \midrule
    Score SDE \citep{song2020score} & 2.38 & 2000 \\ 
    Score SDE-deep \citep{song2020score} & 2.20 & 2000 \\ 
    EDM \citep{Karras2022edm} & 2.01 & 35 \\ 
    EDM (DPM-Solver-v3) \citep{zheng2024dpm} & 2.51 & 10 \\ 
    \midrule
    {\color{gray} Diffusion Distillation} & ~ & ~ \\ 
    \midrule
    PD \citep{salimans2022progressive} & 8.34 & 1 \\ 
    GET~\citep{geng2024one}  & 5.49 & 1 \\
    Diff-Instruct~\citep{di}  & 4.53 & 1 \\
    TRACT \citep{berthelot2023tract} & 3.32 & 2 \\
    CD (LPIPS) \citep{cm} & 3.55 & 1 \\ 
    CD (LPIPS) \citep{cm} & 2.93 & 2 \\ 
    \midrule
    {\color{gray} Consistency Models} & ~ & ~ \\ 
    \midrule
    iCT~\citep{ict}      & 2.83 & 1 \\ 
                         & 2.46 & 2 \\
    iCT-deep~\citep{ict} & 2.51 & 1 \\ 
                         & 2.24 & 2 \\ 
    \midrule
    {\color{gray} ECT} & ~ & ~ \\ 
    \midrule
    ECM (100k iters) & 4.54 & 1 \\ 
    ECM (200k iters) & 3.86 & 1 \\ 
    ECM (400k iters) & 3.60 & 1 \\ 
    ECM (100k iters) & 2.20 & 2 \\ 
    ECM (200k iters) & 2.15 & 2 \\
    ECM (400k iters) & 2.11 & 2 \\ 
    \bottomrule
    \end{tabular}
\end{adjustbox}
\end{minipage}%
\hfill
\begin{minipage}[t]{0.49\linewidth}
\centering
\setlength{\tabcolsep}{10pt} 
\renewcommand{\arraystretch}{0.97} 
\begin{adjustbox}{max width=\textwidth}
\begin{tabular}{@{}lcc@{}}
\toprule
\multicolumn{3}{c}{\imgnet} \\ 
\midrule
Method & FID$\downarrow$ & NFE$\downarrow$ \\ 
\midrule
{\color{gray} Diffusion Models} & ~ & ~ \\ 
\midrule
ADM~\citep{Karras2022edm} & 2.07 & 250  \\
EDM~\citep{Karras2022edm} & 2.22 & 79  \\
EDM2-XL~\citep{karras2023analyzing} & 1.33 & 63 \\ 
\midrule
{\color{gray} Diffusion Distillation} & ~ & ~ \\ 
\midrule
BOOT~\citep{gu2023boot} & 16.3 & 1 \\
DFNO (LPIPS)~\citep{dfno} & 7.83 & 1 \\ 
Diff-Instruct~\citep{di} & 5.57 & 1 \\
TRACT \citep{berthelot2023tract} & 4.97 & 2  \\  
PD (LPIPS) \citep{salimans2022progressive} & 5.74 & 2 \\ 
CD (LPIPS) \citep{cm} & 4.70 & 2 \\ 
\midrule
{\color{gray} Consistency Models} & ~ & ~ \\ 
\midrule
iCT~\citep{ict}      & 3.20 & 2 \\ 
iCT-deep~\citep{ict} & 2.77 & 2 \\
\midrule
{\color{gray} ECT} & ~ & ~ \\ 
\midrule
ECM-S (100k iters) & 3.18 & 2 \\
ECM-M (100k iters) & 2.35 & 2 \\
ECM-L (100k iters) & 2.14 & 2 \\
ECM-XL (100k iters) & 1.96 & 2 \\
\midrule
ECM-S$^\star$  & 4.05  & 1 \\
ECM-S$^\star$  & 2.79  & 2 \\
ECM-XL$^\star$ & 2.49  & 1 \\
ECM-XL$^\star$ & 1.67  & 2 \\
\bottomrule
\end{tabular}
\end{adjustbox}
\end{minipage}

\label{table:ect-cifar10-imagenet64}
\vspace{-10pt}
\end{table}

% ECT unlocks state-of-the-art few-step generative abilities through a simple yet principled approach.
% With a negligible tuning cost, ECT demonstrates strong results while benefiting from the scaling in training FLOPs to enhance its few-step generation capability. 

\subsection{Scaling Laws of ECT}
\begin{figure}[t!]
    \centering
    \includegraphics[width=0.48\linewidth]{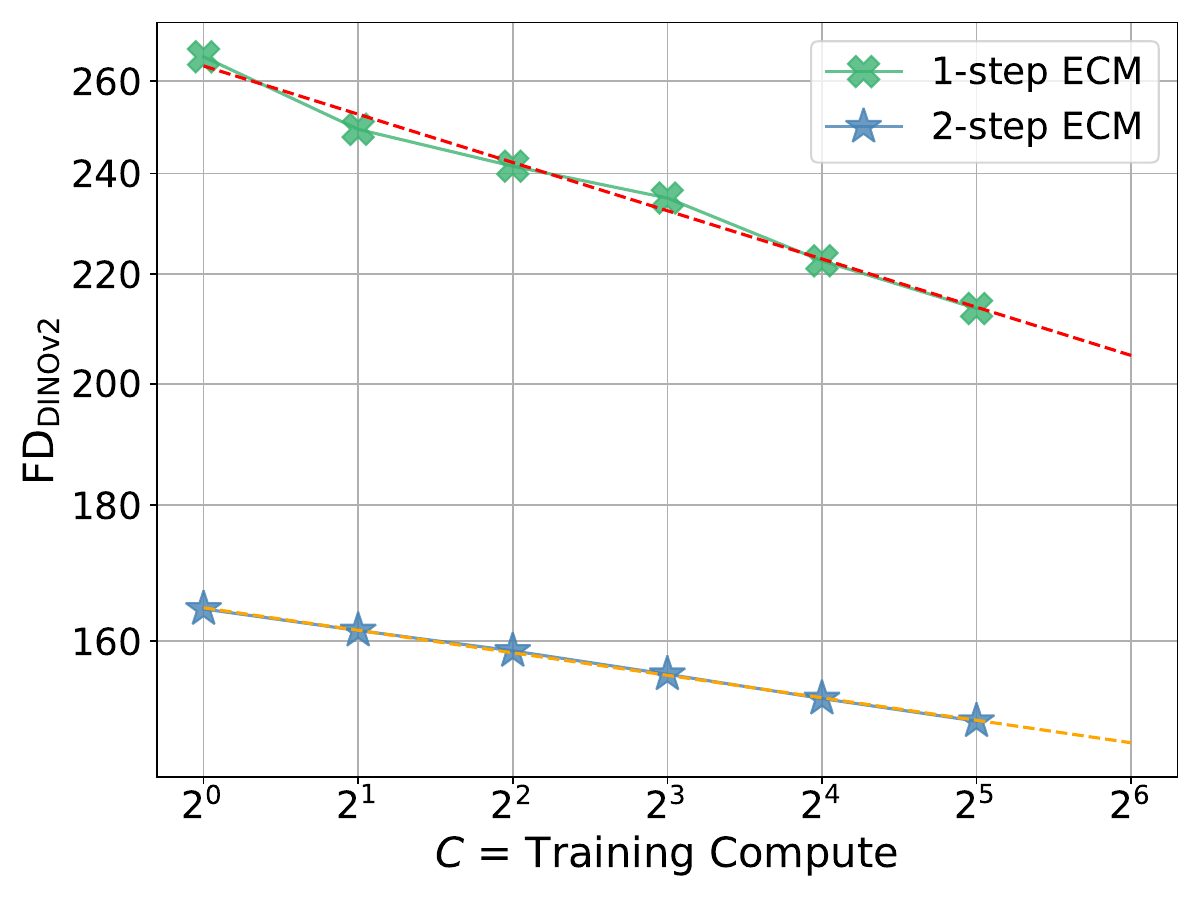}
    \hfill
    \includegraphics[width=0.48\linewidth]{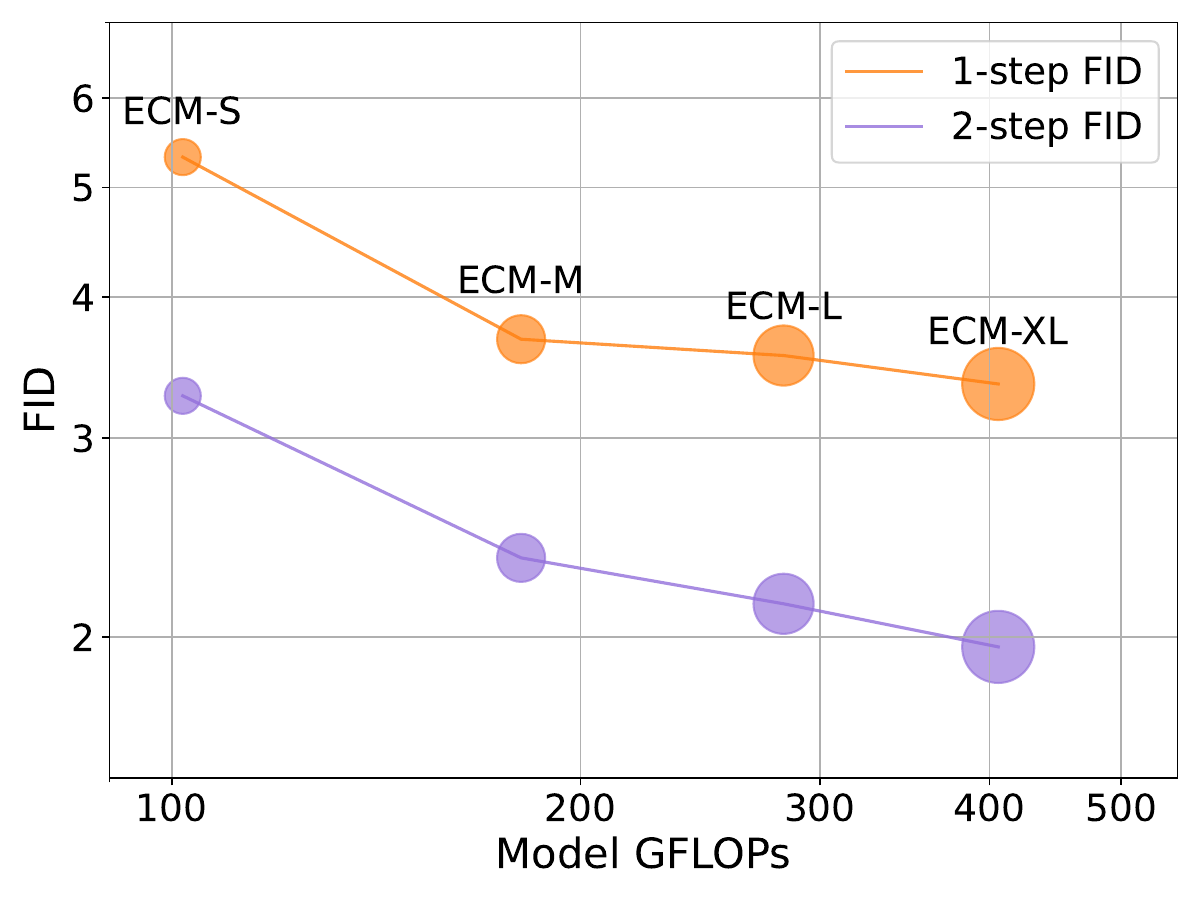}
    \vspace{-5pt}
    \caption{(Left): Scaling up training compute yields the classic power-law between $\mathrm{FD}_{\text{DINOv2}}\downarrow$ and training compute, with $K=263$, $\alpha=-0.060$ for 1-step inference, and $K=164$, $\alpha=-0.028$ for 2-step inference,
    (Right): Given the same batch size and iterations, scaling up model sizes and model FLOPs strongly correlates with $\mathrm{FID}\downarrow$ improvements on ImageNet $64 \times 64$. The diameter is proportional to the model size.}
    \label{fig:imgnet-scaling}
    \vspace{-10pt}
\end{figure}

We leverage efficiency of ECT to examine the scaling behavior of CMs, including training compute, model size, and model FLOPs.
We find that when computational resources are not a bottleneck, ECT scales well and follows the classic power law.
\vspace{-5pt}
\paragraph{Training Compute.} 
Initializing from the weights of EDM~\citep{Karras2022edm}, we fine-tune ECMs across six compute scales on CIFAR-10~\citep{cifar} and plot the trend of \dino\ against training compute in \cref{fig:imgnet-scaling} (Left). The largest compute reaches $2\times$ the diffusion pretraining budget. As we scale up the training budget, we observe a classic power-law decay in \dino, indicating that increased computational investment in ECT leads to substantial improvements in generative performance. Intriguingly, the gap between 1-step and 2-step generation becomes \textit{narrower} when scaling up training compute, even while using the same $\Delta t \to \mathrm{d}t$ schedule. We further fit the power-law $\text{FD}_{\text{DINOv2}} = K \cdot C^\alpha$, where $C$ is the normalized training FLOPs. The Pearson correlation coefficient between $\log(\text{Training Compute})$ and $\log(\text{FD}_{\text{DINOv2}})$ for 1-step and 2-step generation is $-0.9940$ and $-0.9996$, respectively, both with statistical significance ($p$-values $< 10^{-4}$).
\vspace{-5pt}
\paragraph{Model Size \& FLOPs.} Initialized from EDM2~\citep{Karras2024edm2} pretraining, we train ECM-S/M/L/XL models with parameters from $280\mathrm{M}$ to $1.1\mathrm{B}$ and model FLOPs from $102\mathrm{G}$ to $406\mathrm{G}$. As demonstrated in \cref{fig:imgnet-scaling}, both 1-step and 2-step generation capabilities exhibit log-linear scaling for model FLOPs and parameters. This scaling behavior confirms that ECT effectively leverages increased model sizes and computational power to improve 1-step and 2-step generative capabilities. 

Notably, ECT achieves \textit{better} 2-step generation performance than \sota CMs, while utilizing only \textcolor{tomato}{$\mathbf{33\%}$} of the overall computational budget compared to iCT~\citep{ict} ($\text{batch size } 4096 \times 800\text{k}$). This significant efficiency is achieved through a two-stage process: pretraining and tuning. While the pretraining stage utilizes the EDM2 pipeline, the tuning stage of ECT requires a remarkably modest budget of $12.8\mathrm{M}$ training images ($\text{batch size } 128 \times 100\text{k}$), ranging from \textcolor{tomato}{$\mathbf{0.60\%}$ to $\mathbf{1.91\%}$} of the pretraining budget, depending on the model sizes. 
\vspace{-5pt}
\paragraph{Inference.} Our scaling study also indicates a sweet spot for the inference of CMs. On both CIFAR-10 and \imgnet, there are 2-step inferences of smaller models surpassing 1-step inferences of larger models, \textit{e.g.}, $498\mathrm{M}$ ECM-M against $1.1\mathrm{B}$ ECM-XL. This calls for further studies of inference-optimal scaling and test-time compute scaling for visual generation.

\section{Related Work}

\paragraph{Consistency Models.} Consistency models~\citep{cm, ict} are a new family of generative models designed for efficient generation with few model steps without the need for adversarial training. CMs do not rely on a pretrained diffusion model (DM) to generate training targets but instead leverage an unbiased score estimator.
Training CMs have been extended to multi-step sampling~\citep{kim2024consistency,wang2024phased,heek2024multistep}, latent space models~\citep{luo2023latent}, ControlNet~\citep{xiao2023ccm}, video~\citep{wang2024animatelcm}, and combined with an additional adversarial loss~\citep{kim2024consistency,kong2023act}.
Despite their sampling efficiency, CMs are typically more challenging to train and require significantly more compute resources than their diffusion counterparts. Our work substantially improves the training efficiency of CMs, reducing the cost of future research and deployment on CMs.
Parallel to our work, \citet{heek2024multistep,wang2024phased} relax the single-step constraint of CMs and generalize them to multi-step variants, drawing inspiration from TRACT~\citep{berthelot2023tract}.
Under the piecewise linear design, increasing the number of segments will also lead to diffusion pretraining, showing a different approach to connect CMs to the standard diffusion models.

\paragraph{Diffusion Distillation.} 
Drawing inspiration from knowledge distillation~\citep{hinton2015distilling}, distillation is the most widespread training-based approach to accelerate the diffusion sampling procedure. In diffusion distillation, a pretrained diffusion model (DM), which requires hundreds to thousands of model evaluations to generate samples, acts as a teacher. A student model is trained to match the teacher model's sample quality, enabling it to generate high-quality samples in a few steps.

There are two main lines of work in this area. The first category involves trajectory matching, where the student learns to match points on the teacher's sampling trajectory. Methods in this category include 
offline distillation~\citep{Luhman2021KnowledgeDI,dfno,geng2024one}, which require an offline synthetic dataset generated by sampling from a pretrained DM to distill a teacher model into a few-step student model; 
progressive distillation~\citep{salimans2022progressive,meng2023distillation}, and TRACT~\citep{berthelot2023tract}, which require multiple training passes or offline datasets to achieve the same goal; 
and BOOT~\citep{gu2023boot}, Consistency Distillation (CD)\citep{cm}, and Imagine-Flash~\citep{kohler2024imagine}, which minimize the difference between the student predictions at carefully selected points on the sampling trajectory.

CD is closely related to our method, as it leverages a teacher model to generate pairs of adjacent points and enforces the student predictions at these points to map to the initial data point. However, it employs a fixed schedule derived from a specific sampler, which may introduce non-negligible discretization errors in approximating the consistency condition. It also limits the quality of consistency models to that of the pretrained diffusion model. 
% The LPIPS-based metric~\citep{lpips} used in CD loss introduces undesirable bias when evaluated on metrics such as FID.

The second category minimizes the probabilistic divergence between data and model distributions, \ie distribution matching~\citep{poole2022dreamfusion,wang2024prolificdreamer,di,yin2023one,sid}. These methods~\citep{di,yin2023one,nguyen2023swiftbrush,sauer2023adversarial,kohler2024imagine,xu2023ufogen,lin2024sdxl,zhou2024long} use score distillation or adversarial loss, to distill an expensive teacher model into an efficient student model. However, they can be challenging to train in a stable manner due to the alternating updating schemes from either adversarial or score distillation. Some of these methods such as DreamFusion~\citep{poole2022dreamfusion} and ProlificDreamer~\citep{wang2024prolificdreamer} are used for 3D object generation.  

A drawback of training-based approaches is that they need additional training procedures after pretraining to distill an efficient student, which can be computationally intensive. For a detailed discussion on the recent progress of diffusion distillation, we direct the readers to \citet{dieleman2024distillation}.

\vspace{-5pt}
\paragraph{Fast Samplers for Diffusion Models.}
Fast samplers are usually training-free and use advanced solvers to simulate the diffusion stochastic differential equation (SDE) or ordinary differential equation (ODE) to reduce the number of sampling steps. These methods reduce the discretization error during sampling by analytically solving a part of SDE or ODE~\citep{lu2022dpm,lu2022dpmplus,xue2024sa}, by using exponential integrators and higher order polynomials for better approximation of the solution~\citep{zhang2022fast}, using higher order numerical methods~\citep{Karras2022edm}, using better approximation of noise levels during sampling~\citep{kong2021fast}, correcting predictions at each step of sampling~\citep{zhao2024unipc} and ensuring that the solution of the ODE lies on a desired manifold~\citep{liu2022pseudo}. 
Another orthogonal strategy is the parallel sampling process~\citep{pokle2022deep,shih2024parallel}, solving fixed points of the entire ODE/SDE trajectories. A drawback of these fast samplers is that the quality of samples drastically reduces as the number of sampling steps falls below a threshold such as 10 steps.

\section{Conclusion}
We propose Easy Consistency Tuning (ECT), a simple yet efficient scheme for training consistency models. 
The resulting models, ECMs, unlock \sota few-step generative capabilities at a minimal tuning cost and are able to benefit from scaling.
We have made our \href{https://github.com/locuslab/ect}{code} available to ease future prototyping, studying, and deploying consistency models within the community.

\section{Limitations}
One of the major limitations of ECT is that it requires a dataset to tune DMs to CMs. Recent works developed data-free approaches~\citep{gu2023boot,di,yin2023one,sid} for diffusion distillation.
The distinction between ECT and data-free methods is that ECT learns the \textit{consistency condition} on a given dataset through the self teacher, while data-free methods transfer knowledge from a frozen diffusion teacher.
This feature of ECT can be a potential limitation since the training data of bespoke models are unavailable to the public. 
However, we hold an optimistic view on tuning CMs using datasets different from pretraining.

In \cref{sec:ecd}, we discuss a data-free variant of Easy Consistency Distillation (ECD), in which we extend the continuous time formulation to Consistency Distillation, by generating self-synthetic data from CM on-the-fly. However, it only applies to the distillation settings.
Further data-centric research is needed regarding 
synthetic data, data composition, and data scaling for consistency models.

\section*{Broader Impacts and Ethics Statement}\label{sec:broader-impact}

We propose Easy Consistency Tuning (ECT) that can efficiently train consistency models as \sota few-step generators, using only a small fraction of the computational requirements compared to current CMs training and diffusion distillation methods. 
We hope that ECT will democratize the creation of high-quality generative models, enabling artists and creators to produce content more efficiently.
While this advancement can aid creative industries by reducing computational costs and speeding up workflows, it also raises concerns about the potential misuse of generative models to produce misleading, fake, or biased content. 
We conduct experiments on academic benchmarks, whose resulting models are less likely to be misused. Further experiments are needed to better understand these consistency model limitations and propose solutions to address them.

\section*{Reproducibility Statement}\label{sec:reproducibility}
We provide extensive details of experimental settings and hyperparameters to reproduce our experimental results in \cref{sec:experiment-details}. We have provided a zip file of our source code in this submission. We plan to release our code to ensure transparency and reproducibility of the results.

\section*{Acknowledgments}
Zhengyang Geng and Ashwini Pokle are supported by funding from the Bosch Center for AI. Zico Kolter gratefully acknowledges Bosch's funding for the lab. The FLAME Center at CMU generously supports this research.

We appreciate Yang Song for discussions and feedback. We acknowledge the discussions and comments provided by Tianle Cai, Antony Jia, Diyang Xue, Runtian Zhai, and Yonghao Zhuang during the preparation of the \href{https://gsunshine.notion.site/Consistency-Models-Made-Easy-954205c0b4a24c009f78719f43b419cc}{blog post}.

% \clearpage
\bibliography{ref}
\bibliographystyle{unsrtnat} 

\newpage
\appendix

\section{Motivations behind Design Choices in ECT}
\label{sec:design-choices}

In this section, we expand upon our motivation behind the design decisions for the mapping function, metric, and weighting function used for ECT.

\paragraph{Mapping Function.}
We first assume that $\Delta t$ is \textit{approximately} proportional to $t$.
Let $0 < c \leq 1$ be this constant of proportionality, then we can write:
\begin{align*}
&c \approx \frac{\Delta t}{t} = \frac{t - r}{t} = 1 - \frac{r}{t} \Rightarrow
\frac{r}{t} \approx 1 - c.
\end{align*}

As training progresses, the mapping function should gradually shrink $\Delta t \to 0$. However, the above parameterization does not achieve this. An alternative parameterization is to decrease $\Delta t$ exponentially.
We assume the ratio between $r$ and $t$ can be written as:
\begin{equation}
\label{eq:const-mapping}
\frac{r}{t} = 1 - \frac{1}{q^a},
\end{equation}
where $q > 1$, $a = \lfloor\text{iters}/d\rfloor$, and $d$ is a hyperparameter controlling how quickly $\Delta t \to \d t$. At the beginning of training, $\frac{r}{t} = 1 - \frac{1}{q^0} = 0 \Rightarrow r = 0$, which falls back to DMs. Since we can initialize from the diffusion pretraining, this stage can be skipped by setting $a = \lceil\text{iters}/d\rceil$. As training progresses ($\text{iters} \uparrow$), $\frac{r}{t} \to 1$ leads to $\Delta t \to \d t$.

Finally, we adjust the mapping function to balance the prediction difficulties across different noise levels:
\begin{equation}
\label{eq:adj-mapping}
\frac{r}{t}
= 1 - \frac{1}{q^a} n(t)
= 1 - \frac{1}{q^{\lceil\text{iters}/d\rceil}} n(t).
\end{equation}
For $n(t)$, we choose $n(t) = 1 + k ,\sigma(-b,t) = 1 + \frac{k}{1 + e^{bt}}$, using the sigmoid function $\sigma$. Since $r\geq0$, we also clamp $r$ to satisfy this constraint after the adjustment.

\begin{figure}[htbp!]
    \vspace{10pt}
    \centering    
    \includegraphics[width=0.8\linewidth]{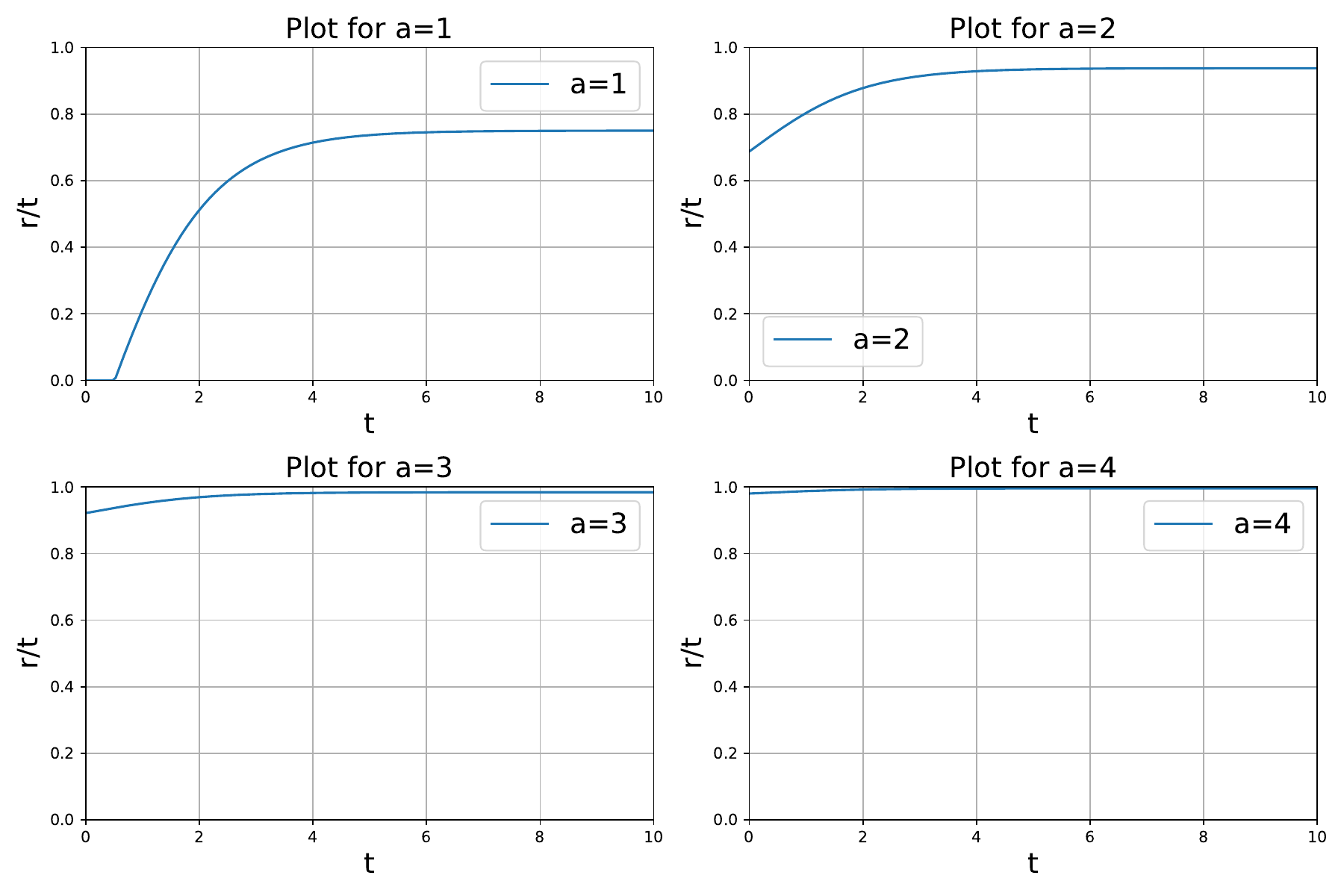}
    \caption{Visualization of $\nicefrac{r}{t}$ during training. $\Delta t \to \d t$ when $\nicefrac{r}{t} \to 1$.}
    \label{fig:mapping}
\end{figure}
\vspace{0.3cm}

The intuition behind this mapping function is that the relative difficulty of predicting $f(\x_r)$ from $\x_t$ can vary significantly across different noise levels $t$ when using a linear mapping between $t$ and $r$.

Consider $\nicefrac{r}{t} = 0.9$. At small values of $t$, $\x_t$ and $\x_r$ are close, making the alignment of $f(\x_t)$ with $f(\x_r)$ relatively easy. In contrast, at larger $t$, where $\x_t$ and $\x_r$ are relatively far apart, the distance between the predictions $f(\x_t)$ and $f(\x_r)$ can be substantial. This leads to imbalanced gradient flows across different noise levels, impeding the training dynamics.

Therefore, we downscale $\nicefrac{r}{t}$ when $t$ is near $0$ through the mapping function, balancing the gradient flow across varying noise levels. 
This prevents the gradient at any noise level from being too small or too large relative to other noise levels, thereby controlling the variance of the gradients. 

We direct the reader to \cref{sec:additional-experimental-results} for details of how to set $q^{\lceil\text{iters}/d\rceil}$.

\paragraph{Choice of Metric.}
As discussed in \cref{sec:preliminaries}, iCT uses pseudo-Huber metric~\citep{huber} to mitigate the perceptual bias caused by the LPIPS metric~\citep{lpips}, 
\begin{align}
L(\x, \y) = \sqrt{\| \x - \y \|_2^2 + \epsilon} - \epsilon, \quad \epsilon > 0.
\end{align}
This metric indeed improves the performance of CMs over the classic squared $L_2$ loss. When taking a careful look as this metric, we reveal that one of the reasons for this improvement is that this metric is more robust to the outliers compared to the $L_2$ metric due to its adaptive per-sample scaling of the gradients.
Let $\Delta = \x - \y$, then the differential of the pseudo-Huber metric can be written as
\begin{equation}
\label{eq:grad-huber}
\mathrm{d} L = \underbrace{\frac{1}{\sqrt{\|\Delta\|_2^2 + c^2}} \vphantom{\mathrm{d} \left( \frac{1}{2} \|\Delta\|_2^2 \right)} }_{\text{weighting term}} \underbrace{\mathrm{d} \left( \frac{1}{2} \|\Delta\|_2^2 \vphantom{\frac{1}{\sqrt{\|\Delta\|_2^2 + \epsilon}}}\right)}_{\text{differential of 
 squared } L_2 \text{\,loss}},
\end{equation}
where we have decomposed the differential of pseudo-Huber loss into an adaptive weighting term and the differential of the squared $L_2$ loss.
Therefore, we retain the squared $L_2$ metric used in DMs, and explore varying adaptive weighting terms which we explore in detail in \cref{sec:additional-experimental-results}.

\paragraph{Distinction between the training schedules of ECT and iCT.} 
As noted in \cref{sec:preliminaries}, iCT \citep{ict} employs a discrete-time curriculum given by ~\cref{schedule}. 
This curriculum divides the noise horizon $[0,T]$ into $N$ smaller consecutive subintervals to apply the consistency loss, characterized by non-overlapping segments $[t_i, t_{i+1}]$, and gradually increases the number of intervals $N=10 \to 1280$. 
However, the "boundary" condition of this schedule is to start with the number of intervals to $N=1$, learning a model solely mapping samples at noise levels $T_{\text{max}}$ to the clean data $\x_0$, largely distinct from the classic diffusion models training.
We instead investigate a continuous-time schedule whose "boundary" condition yields diffusion pretraining, \ie constructing training pairs of $r=0$ for all $t$ at the beginning.

% We propose a vastly different continuous-time training schedule that aids in mitigating the "curse of consistency" as discussed in \cref{sec:tacking-curse-of-consistency}.

\section{Exploring Design Space \& Scaling of Consistency Models}
\label{sec:additional-experimental-results}

Due to ECT's efficiency, we can explore the design space of CMs at a minimal cost. We specifically examine the weighting function, training schedule, and regularization for CMs.

Our most significant finding is that \textit{controlling gradient variances and balancing the gradients across different noise levels are fundamental to CMs' training dynamics}. Leveraging the deep connection between CMs and DMs, we also improve the diffusion pretraining and the full pretraining+tuning pipeline using our findings.

\paragraph{Weighting Function.} Forward processes with different noise schedules and model parameterizations can be translated into each other at the cost of varying weighting functions~\citep{wt}.
From our experiments on a wide range of weighting schemes, we learn three key lessons.

\begin{table}[t!]
\vspace{5pt}
\caption{Performance of ECMs trained with various weighting functions on \imgnet. We enable the adaptive weighting $w(\Delta) = \nicefrac{1}{(\|\Delta\|_2^2 + c^2)^{\frac{1}{2}}}$.}
\vspace{5pt}
\centering
\begin{tabular}{@{}lcc@{}} 
\toprule
$\Bar{w}(t)$ & 1-step FID$\downarrow$ & 2-step FID$\downarrow$ \\ 
\midrule
$1$                   & \textbf{5.39}   & 3.48  \\
$\nicefrac{1}{t}$     & 17.79	        & 3.24  \\
$\nicefrac{1}{(t-r)}$ & 9.28	        & 3.22  \\
$\nicefrac{1}{t}+\nicefrac{1}{\sigma_{\mathrm{data}}}$  & 5.68 & 3.44  \\
$\nicefrac{1}{t^2}$   & 190.80          & 20.65         \\
$\nicefrac{1}{t^2}+1$ & 6.78            & \textbf{3.12} \\
$\nicefrac{1}{t^2}+\nicefrac{1}{\sigma_{\mathrm{data}}^2}$ & 5.51 & 3.18 \\
$\nicefrac{1}{(t^2 +\sigma_{\mathrm{data}}^2)}$ & 163.01          & 13.33 \\
\bottomrule
\end{tabular}
\label{tab:wt-imgnet}
\end{table}

$(1)$ \textit{There is no free lunch for weighting function}, \ie there is likely no universal timestep weighting $\Bar{w}(t)$ that can outperform all other candidates on different datasets, models, and target metrics for both 1-step and 2-step generation. 

We refer these results to \cref{tab:wt-imgnet}, including $\mathrm{SNR}(t) = \nicefrac{1}{t^2}$, $\mathrm{SNR}(t)+1 = \nicefrac{1}{t^2}+1$~\citep{salimans2022progressive}, EDM weighting $\mathrm{SNR}(t) + \nicefrac{1}{\sigma_{\mathrm{data}}^2} = \nicefrac{1}{t^2}+\nicefrac{1}{\sigma_{\mathrm{data}}^2}$~\citep{Karras2022edm}, and Soft-Min-SNR weighting $\nicefrac{1}{(t^2 +\sigma_{\mathrm{data}}^2)}$~\citep{min-snr,soft_min_snr}, where $\mathrm{SNR}(t) = \nicefrac{1}{t^2}$ is the signal-to-noise ratio in our setup.

On CIFAR-10, the weighting $\Bar{w}(t) = \nicefrac{1}{(t-r)}$ from the discretization of consistency condition in \cref{eq:CMcost} achieves the best 1-step FID, while the square root of $\mathrm{SNR}(t)$, $\Bar{w}(t) = \sqrt{\mathrm{SNR}(t)} = \nicefrac{1}{t}$, produces the best \dino. 
On \imgnet, considering that we have already had the adaptive weighting $w(\Delta)$, the uniform weighting $\Bar{w}(t) \equiv 1$ can demonstrate the best 1-step FID when tuning from EDM2~\citep{Karras2024edm2}.
In contrast to 1-step FIDs, a wider range of timestep weighting $\Bar{w}(t)$ produces close 2-step FIDs for ECMs. 

When starting on a new dataset with no prior information, $\Bar{w}(t) = \mathrm{SNR}(t)+n$ is a generally strong choice as the default timestep weighting of data prediction models (x-pred). In this situation, this weighting function corresponds to using v-pred~\citep{salimans2022progressive} or flow matching~\citep{lipman2022flow,liu2022flow} as model parameterization when $n=1$.

\begin{table}[t!]
\caption{Performance of ECMs trained with varying adaptive weightings on \imgnet.}
\vspace{5pt}
\centering
\begin{tabular}{@{}llcc@{}} 
\toprule
$\Bar{w}(t)$ & $w(\Delta)$ & 1-step FID$\downarrow$ & 2-step FID$\downarrow$ \\ 
\midrule
$\nicefrac{1}{t^2}+\nicefrac{1}{\sigma_{\mathrm{data}}^2}$ & $1$ & 6.51   & 3.28     \\
$\nicefrac{1}{t^2}+\nicefrac{1}{\sigma_{\mathrm{data}}^2}$ & $\nicefrac{1}{(\|\Delta\|_1 + c})$          & 6.29	 & 3.25	 \\
$\nicefrac{1}{t^2}+\nicefrac{1}{\sigma_{\mathrm{data}}^2}$ &  $\nicefrac{1}{(\|\Delta\|_2^2 + c^2)^{\frac{1}{2}}}$ & 5.51   & 3.18     \\
$1$ & $\nicefrac{1}{(\|\Delta\|_2^2 + c^2)^{\frac{1}{2}}}$ &  5.39  & 3.48 \\
\bottomrule
\end{tabular}
\label{tab:adaptive-wt-imgnet}
\end{table}
\vspace{0.2cm}

$(2)$ \textit{The adaptive weighting $w(\Delta)$ achieves better results by controlling gradient variance}. The adaptive weighting $w(\Delta)$ on a per-sample basis shows uniform improvements on both CIFAR-10 and \imgnet. See \cref{tab:adaptive-wt-imgnet} for the ablation study. 

Beyond ECT, we further investigate the role of adaptive weighting \( w(\Delta) \) in pretraining on a toy Swiss roll dataset using the parameterization and forward process of flow matching~\citep{lipman2022flow} and an MLP network.

Consider the objective function \( w(\Delta) \|v_\theta(\x_t) - (\x_1 - \x_0) \|_2^2 \), where \( \x_t = (1-t) \cdot \x_0 + t \cdot \x_1 \), \(t \sim \mathrm{Uniform}(0, 1)\), \( \x_1 \sim \mathcal{N}(\mathbf{0}, \mathbf{I})\), and the adaptive weighting
\[
w(\Delta) = \frac{1}{(\|\Delta\|_2^2 + \epsilon)^p},
\]
where \( p=0 \) corresponds to no adaptive weighting. We set \( \epsilon = 10^{-6} \) and control the strength of gradient normalization by varying \( p \) from $0$ to $1$.

\begin{figure}[t!]
    \vspace{10pt}
    \centering 
    \includegraphics[width=0.8\linewidth]{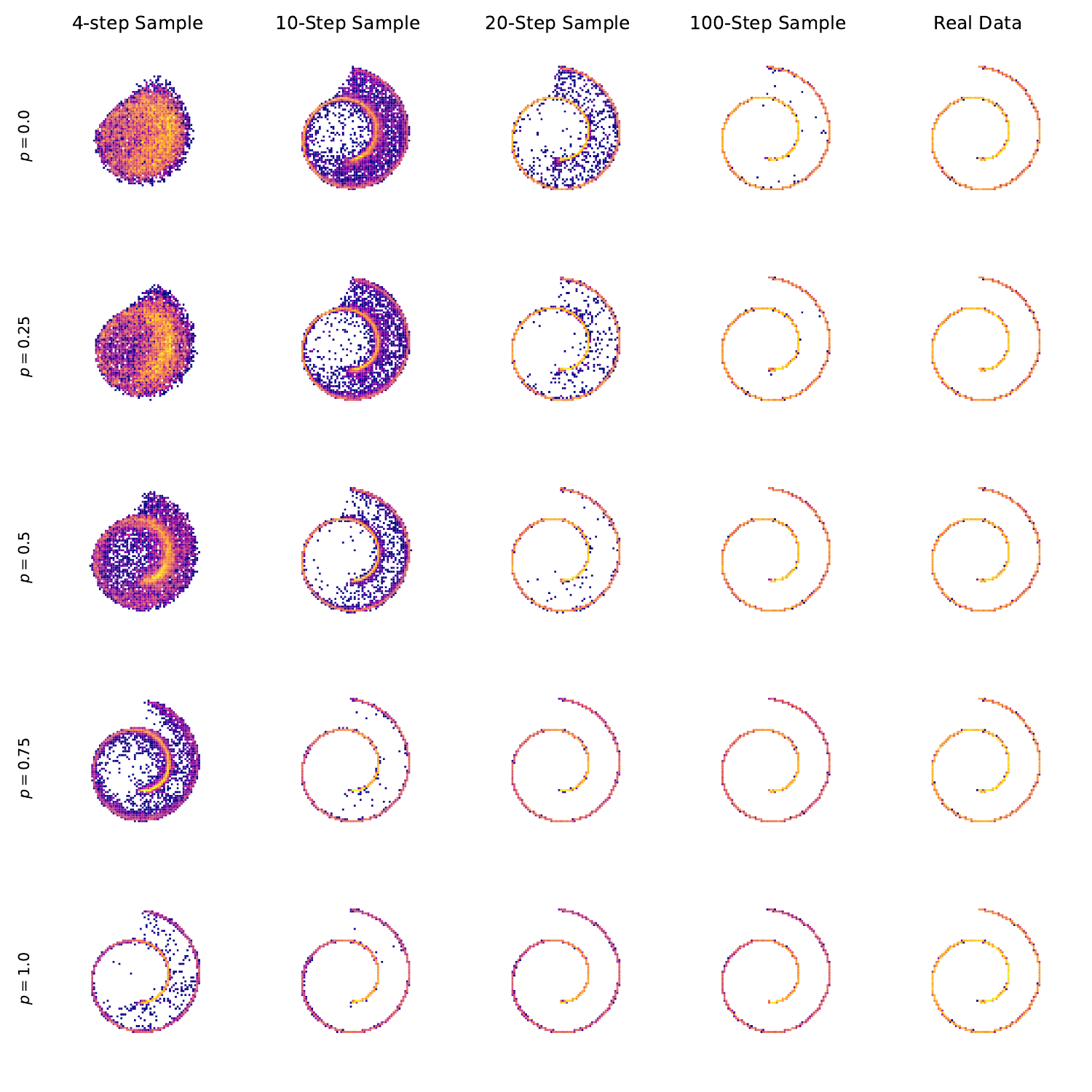}
    \caption{Influence of adaptive weighting $w(\Delta) = \nicefrac{1}{(\|\Delta\|_2^2 + \epsilon)^p}$ on pretraining using varying $p$.}
    \label{fig:adp-wt}
    \vspace{-10pt}
\end{figure}

As we increase the strength of adaptive weighting, flow models become easier to sample from in a few steps. Surprisingly, even \( p=1 \) demonstrates strong few-step sampling results when pretraining the flow model. See \cref{fig:adp-wt} for visualization.

% $(3)$ \textit{The timestep weighting $\Bar{w}(t)$ and the adaptive weighting $w(\Delta)$ are compatible}. This compatibility is how we build ECMs in this work, using adaptive weighting \( w(\Delta) \) to achieve strong 1-step sampling capabilities and timestep weighting \( \Bar{w}(t) \) to improve 2-step sampling.

\paragraph{Mapping Function.} We compare the constant mapping function with $n(t) \equiv 1$ in \cref{eq:const-mapping} and mapping function equipped with the sigmoid $n(t)$ in \cref{eq:adj-mapping}. We use $k=8$ and $b=1$ for all the experiments, which transfers well from CIFAR-10 to \imgnet\ and serves as a baseline in our experiments. Though $b=2$ can further improve the 1-step FIDs on \imgnet, noticed post hoc, we don't rerun our experiments.

On CIFAR-10, the constant mapping function with $n(t) \equiv 1$ achieves 1-step FID of 4.06 at $200$k iterations, worse than the 1-step FID of 3.86 by $n(t)= 1 + \frac{k}{1 + e^{bt}}$.
Under our forward process ($\x_t = \x_0 + t \cdot \epsilon$) and model parameterization (EDM~\citep{Karras2022edm}), the constant mapping function incurs training instability on \imgnet, likely due to the imbalanced gradient flow. 

The role of the mapping function, regarding training, is to balance the difficulty of learning \textit{consistency condition} across different noise levels, avoiding trivial consistency loss near $t \to 0$.
For model parameterizations and forward processes different from ours, for example, flow matching~\citep{lipman2022flow,liu2022flow,sd3}, we advise readers to start from the constant mapping function due to its simplicity.

\vspace{-10pt}
\paragraph{Dropout.} In line with \citep{ict}, we find that CMs benefit significantly from dropout~\citep{dropout}.  
On CIFAR-10, we apply a dropout of $0.20$ for models evaluated on FID and a dropout of $0.30$ for models evaluated on \dino. 

On \imgnet, we note that ECT benefits from a surprisingly high dropout rate. When increasing the dropout rate from $0.10$ to $0.40$, the 2-step FID decreases from $4.53$ to $3.24$. 
Increasing the dropout rate further can be helpful for 1-step FID under certain timestep weighting $\Bar{w}(t)$, but the 2-step FID starts to deteriorate.
In general, we optimize our model configurations for 2-step generation and choose the dropout rate of $0.40$ for ECM-S.

Finally, we note that the dropout rate tuned at a given weighting function $w(t)$ transfers well to the other weighting functions, thereby reducing the overall cost of hyperparameter tuning. 
On \imgnet, the dropout rate can even transfer to different model sizes. We apply a dropout rate of $0.50$ for all the model sizes of ECM-M/L/XL.

\paragraph{Shrinking $\Delta t \to \mathrm{d}t$.} 
In the mapping function discussed in \cref{sec:ect} and \cref{sec:design-choices}, we use the hyperparameter $d$ to control the magnitude of $q$, thereby determining the overall rate of shrinking $\Delta t \to \mathrm{d}t$, given by $\left(1 - \nicefrac{1}{q}^{\lceil\text{iters}/d\rceil} \right)$.
In practice, we set $q=2$ and $d = \text{total\_iters}//8$ for CIFAR-10 experiments, and $q=4$ and $d = \text{total\_iters}//4$ for \imgnet\ experiments, achieving $\nicefrac{r}{t} \approx 0.99$ at the end of training.

Compared with no shrinkage of $\Delta t$, where $\Delta t \approx \mathrm{d}t$ throughout, we find that shrinking $\Delta t \to \mathrm{d}t$ results in improved performance for ECMs. 
For example, on CIFAR-10, starting ECT directly with $\Delta t \approx \mathrm{d}t$ by setting $q=256$ (corresponding to $\nicefrac{r}{t} \approx 0.99$) leads to quick improvements in sample quality initially but slower convergence later on. The 1-step FID drops from 3.60 to 3.86 using the same 400k training iterations compared to gradually shrinking $\Delta t \to \mathrm{d}t$. 
On \imgnet, $\Delta t \approx \mathrm{d}t$ with $q=256$ from the beginning results in training divergence, as the gradient flow is highly imbalanced across noise levels, even when initializing from pretrained diffusion models.

This observation suggests that \emph{ECT's schedule should be adjusted according to the compute budget}. At small compute budgets, as long as training stability permits, directly approximating the \textit{differential consistency condition} through a small $\Delta t \approx \mathrm{d}t$ leads to fast sample quality improvements. For normal to rich compute budgets, shrinking $\Delta t \to \mathrm{d}t$ generally improves the final sample quality, which is the recommended practice.

Using this feature of ECT, we demonstrate its efficiency by training ECMs to surpass previous Consistency Distillation, which took hundreds of GPU hours, using \textcolor{tomato}{\textbf{one hour on a single A100 GPU}}.

\paragraph{Training Generative Models in 1 GPU Hour.} Deep generative models are typically computationally expensive to train. Unlike training a classifier on CIFAR-10, which usually completes within one GPU hour, leading generative models on CIFAR-10 as of 2024 require days to a week to train on 8 GPUs. Even distillation from pretrained diffusion models can take over a day on 8 GPUs or even more, equivalently hundreds of GPU hours.

To demonstrate the efficiency of ECT and facilitate future studies, we implemented a fast prototyping setting designed to yield results within one hour on a single GPU. This configuration uses a fixed $\Delta t \approx \mathrm{d}t$ by setting $q=256$ in our mapping function (corresponding to $\frac{r}{t} \approx 0.99$), which allows for quick approximation of the differential consistency condition.  Through $8000$ gradient descent steps at batch size of $128$, within 1 hour on a single A100 40GB GPU, ECT achieves a 2-step FID of $2.73$, outperforming Consistency Distillation (2-step FID of $2.93$) trained with 800k iters at batch size 512 and LPIPS~\citep{lpips} metric.

% \section{Ablation Study}

\paragraph{Ablation on Pretraining}
We conducted a controlled experiment, training CMs from scratch following iCT best practices and tuning EDM using both iCT and ECT schedules. The combined pretraining and fine-tuning cost for ECT is $50\% + 6.25\%$ of iCT trained from scratch. Results are presented in \cref{tab:pretraining-impact}.

There is a noticeable performance gap between iCT's training and ECT's pretraining+tuning scheme, particularly when evaluating models using modern metrics like $\mathrm{FD}_{\text{DINOv2}}$~\citep{fd_dino}.  $\mathrm{FD}_{\text{DINOv2}}$ employs the representation space of the DINOv2-L model~\citep{oquab2023dinov2} to compute distributional distance, which has been shown to better align with human evaluations.

\begin{table}[h]
\centering
\caption{Impact of pretraining on model performance}
\vspace{5pt}
\label{tab:pretraining-impact}
\begin{tabular}{lc}
\toprule
Model & $\mathrm{FD}_{\text{DINOv2}}$ \\
\midrule
iCT                       & 242.30 \\
Pretraining + iCT tuning  & 200.31 \\
Pretraining + ECT tuning  & 190.13 \\
EDM                       & 168.16 \\
\bottomrule
\end{tabular}
\vspace{10pt}
\end{table}

When scaling up the pretraining+tuning cost to match the overall cost of iCT in class-conditional settings, ECM achieves a $\mathrm{FD}_{\text{DINOv2}}$ of 152.21, significantly outperforming the iCT model trained from scratch (205.11). For context, StyleGAN-XL achieves an $\mathrm{FD}_{\text{DINOv2}}$ of 204.60.

\begin{table}[t!]
\caption{Generative performance on class-conditional CIFAR-10. }
\vspace{5pt}
\centering
\small
\centering
\setlength{\tabcolsep}{5pt} % Adjust horizontal padding between columns
\renewcommand{\arraystretch}{0.98} % 
\begin{tabular}{@{}lccc@{}}
\toprule
Method & \dino$\downarrow$ & NFE$\downarrow$ \\ 
\midrule
{\color{gray} GANs} & ~ & ~ \\ 
\midrule
BigGAN~\citep{biggan}               & 326.66 & 1 \\
StyleGAN2-ADA~\citep{Karras2020ada} & 305.92 & 1 \\
StyleGAN-XL~\citep{styleganxl}      & 204.60 & 1 \\
\midrule
{\color{gray} Diffusion Models} & ~ & ~ \\ 
\midrule
EDM~\citep{Karras2022edm}	& 145.20 & 35 \\
PFGM++~\citep{xu2023pfgm++}  & 141.65 & 35 \\
\midrule
{\color{gray} ECT} & ~ & ~ \\ 
\midrule
ECM (ECT Pretrained) & 121.05 & 35\\
ECM (Tuned) & 198.51 & 1 \\
ECM	(Tuned) & 128.63 & 2 \\
\bottomrule
\end{tabular}
\label{tab:cifar-dino}
\end{table}

\paragraph{Ablation on Tuning Design Choices}
Besides quantitatively evaluating the pretraining impact, we conduct further ablation studies regarding fine-tuning design choices. Due to the computational constraints, we tune EDM2-S using iCT's design choices as the baseline.  Results are presented in \cref{tab:tuning-ablation}. These results demonstrate the improvements of the continuous-time training schedule, increased dropout, and weighting functions on ECMs' generative performance.

\begin{table}[h]
\centering
\caption{Ablation study of continuous-time CMs on \imgnet}
\vspace{5pt}
\label{tab:tuning-ablation}
\begin{tabular}{lcc}
\toprule
Methods & 1-step FID & 2-step FID \\
\midrule
iCT + EDM2 Pretraining                             & 21.09 & 4.39 \\
+ Continuous-time schedule                         & 14.34 & 4.33 \\
+ Dropout = 0.40                                   & 9.28  & 3.22 \\
+ $\Bar{w}(t)=\nicefrac{1}{t^2}+\nicefrac{1}{\sigma_{\mathrm{data}}^2}$ & 5.51  & 3.18 \\
\bottomrule
\end{tabular}
\vspace{5pt}
\end{table}

\paragraph{Improving Pretraining using Findings in Tuning.} The exploration of the design space through the tuning stage as a proxy led to a question: Can the insights gained during tuning be applied to improve the pretraining stage and, consequently, the entire pretraining+tuning pipeline for CMs? The results of our experiments confirmed this hypothesis.

For the largest $\Delta t = t$, ECT falls back to diffusion pretraining with $r=0$ and thus $\f(\x_r) = \x_0$. 
We pretrain EDM~\citep{Karras2022edm} on the CIFAR-10 dataset using the findings in ECT. Instead of using EDM weighting, $\mathrm{SNR}(t) + \nicefrac{1}{\sigma_{\mathrm{data}}^2}$, we enable the adaptive weighting $w(\Delta)$ with $p=\nicefrac{1}{2}$ and smoothing factor $c=0$ and a timestep weighting $\Bar{w}(t) = \nicefrac{1}{t}$.

Compared with the EDM baseline, the recipe from ECT brings a convergence acceleration over $2\times$ regarding \dino, matching EDM's final performance using less than half of the pretraining budget and largely outperforming it at the full pretraining budget. 

EDM pretrained by ECT achieves \dino\ of $150.39$ for unconditional generation and $121.05$ for class-conditional generation, considerably better than the EDM baseline's \dino\ of $168.17$ for unconditional generation and $145.20$ for class-conditional generation, when using the same pretraining budget and inference steps (NFE=35).

\paragraph{Influence of Pretraining Quality.} Using ECT pretrained models (\dino\ of $121.05$) and original EDM~\citep{Karras2022edm} (\dino\ of $145.20$), we investigate the influence of pretraining quality on consistency tuning and resulting ECMs.
Our experiments confirm that better pretraining leads to easier consistency tuning and faster convergence.
At the same budget of $204.8\mathrm{M}$ images, tuning from ECT pretrained models achieves \dino\ of $128.63$, better than \dino\ of $152.21$ from EDM. 

ECM from the ECT pretraining surpasses SoTA GANs in 1 sampling step and advanced DMs in 2 sampling steps, only slightly falling behind our pretrained models and setting up a new SoTA for the modern metric \dino. Results can be found in \cref{tab:cifar-dino}. 

On \imgnet, ECM-M, initialized from EDM2-M~\citep{Karras2024edm2}, deviates from the power law scaling and achieves better generative performance than the log-linear trend. (See \cref{fig:imgnet-scaling}, Right). We speculate that it is due to a higher pretraining budget, in which EDM2-M was pretrained by $2\times$ training images compared with other model sizes (S/L/XL).

\begin{figure}[t!]
    \centering    
    \includegraphics[width=0.5\linewidth]{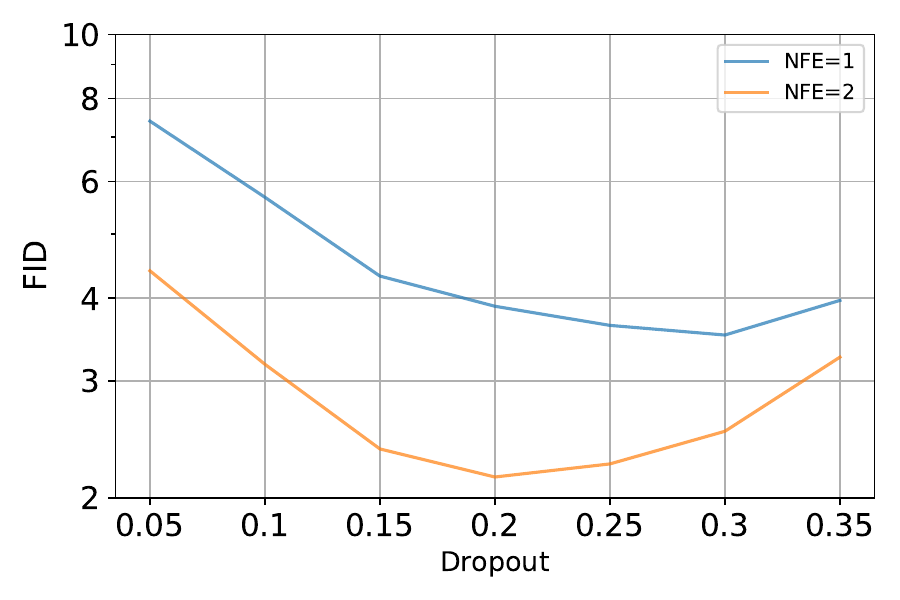}
    \caption{Relationship between the dropout and FIDs for models trained on CIFAR-10 with varying numbers of function evaluations (NFE) at inference.}
    \label{fig:fid-dropout-nfe}
\end{figure}

\paragraph{Differences between 1-step and 2-step Generation.}
Our empirical results suggest that the training recipe for the best 1-step generative models can differ from the best few-step generative models in many aspects, including weighting function, dropout rate, and EMA rate/length.
\cref{fig:fid-dropout-nfe} shows an example of how FIDs from different numbers of function evaluations (NFEs) at inference vary with dropout rates. 

In our setups, starting from a proper model size, the improvements from 2-step sampling seem larger than doubling the model size but keeping 1-step sampling.
In the prior works, iCT~\citep{ict} employs $2\times$ deeper model, but the 1-step generative performance can be inferior to the 2-step results from ECT. 
This finding is consistent with recent theoretical analysis \citep{lyu2023convergence}, which indicates a tighter bound on the sample quality for the 2-step generation compared to the 1-step generation. 

\paragraph{Pareto Frontier \& Scaling Law.}
The Pareto Frontier reveals a seemingly power law scaling behavior. Training configurations not optimized for the current compute budget, \ie not on the Pareto Frontier, deviate from this scaling. Simply scaling up the training compute without adjusting other parameters may result in suboptimal performance. In our compute scaling experiments, we increased the batch size and enabled the smoothing factor $c$ in the adaptive weighting to maintain this trend.

\section{Extension to Consistency Distillation.}
\label{sec:ecd}

\paragraph{Continuous-time Consistency Distillation}
Consistency Distillation (CD)~\citep{cm} uses a fixed discrete schedule derived from a specific sampler, such as the EDM sampler. This approach inherently limits $\Delta t \to \d t$, therefore causing a non-trivial discretization error on approximating the differential consistency condition.  Building upon ECT, we extend our continuous-time schedule to Consistency Distillation, which we term Easy Consistency Distillation (ECD).

On \imgnet, we implement Consistency Distillation as the baseline using pretrained EDM2-S~\citep{Karras2024edm2} with weighting functions, noise distribution, and dropout. The results, shown in \cref{tab:ecd}, demonstrate that continuous-time distillation improves upon the standard CD approach.

Given well-pretrained DMs, ECD typically demonstrates a performance advantage over ECT at smaller batch sizes (e.g., 128), primarily due to the variance reduction in approximating the score function:
\[ \nabla_{\x_{t}} \log p(\x_{t}) = \E [\nabla_{\x_{t}} \log p(\rvx_{t} | \rvx_{0}) \vert \rvx_t ] \]

This advantage stems from the teacher models, which provide a variance-reduced estimate of the score function compared to the Monte Carlo estimation used in ECT. 
However, this gap tends to narrow as we scale up to larger model sizes. 
Scaling compensates for the limited training budget and variances, closing the gap between the two.

\begin{table}[t]
\centering
\caption{Easy Consistency Distillation (ECD) on ImageNet 64$\times$64 using the same budget of $12.8\mathrm{M}$ training images (batch size 128 and 100k iterations) as ECT in \cref{table:ect-cifar10-imagenet64}.}
\vspace{5pt}
\label{tab:ecd}
\begin{tabular}{lcc}
\toprule
Methods & 1-step FID & 2-step FID \\
\midrule
CD-S              & 8.18  & 3.71 \\
ECD-S             & 3.33  & 2.10 \\
ECD-M             & 2.78  & 1.92 \\
ECD-XL            & 2.54  & 1.77 \\
\bottomrule
\end{tabular}
\vspace{-5pt}
\end{table}

Consequently, while ECD may outperform ECT in resource-constrained scenarios or with smaller models, the distinction becomes less pronounced as we move to larger scales. This observation underscores the scaling factor (computational resources and model size) when choosing between ECT and ECD for a given application.

\paragraph{Data-Free ECD.} Inspired by the recent progress on data-free distillation~\citep{gu2023boot,di,yin2023one,sid}, we explore a data-free variant of Easy Consistency Distillation (ECD).
Instead of sampling data points $\x_0$ from a given dataset $\mathcal{D}$, we generate synthetic data points directly from the CM itself on the fly, $\x_0 = f_{\operatorname{sg}(\theta)}(\boldsymbol{\epsilon}', T)$, where $\boldsymbol{\epsilon}’ \sim p(\boldsymbol{\epsilon})$, $T$ is the maximum noise level for sampling, followed by the same ECD training step over the self synthetic data. This eliminates the need for a distillation dataset or the construction of synthetic data from the teacher model~\citep{Luhman2021KnowledgeDI,geng2024one,dfno}.

On \imgnet, this data-free ECD achieves 1-step FID of 4.38 and 2-step FID of 2.77, comparable to the data-dependent ECD and ECT schemes at the same budget. It suggests that data-free ECD can be a competitive alternative in scenarios where access to large datasets is limited or unavailable.
We summarize both ECD and its data-free variant in \cref{alg:ecd}.

\vspace{5pt}
\begin{algorithm}
\caption{Easy Consistency Distillation (ECD) / Data-Free ECD}
\label{alg:ecd}
\begin{algorithmic}[t]
\State \textbf{Input:} Dataset $\gD$ (for ECD), a pretrained diffusion model $\phi$, mapping function $p(r \mid t, \mathrm{Iters})$, weighting function $w(t)$.
\State \textbf{Init:} $\theta \leftarrow \phi$, $\mathrm{Iters} = 0$.
\Repeat
\If{ECD}
\State $\mathrm{Sample}$ $\x_0 \sim \mathcal{D}$
\ElsIf{Data-Free ECD}
\State $\mathrm{Sample}$ $\boldsymbol{\epsilon}' \sim p(\boldsymbol{\epsilon})$
\State $\mathrm{Compute}$ $\x_0 = f_{\operatorname{sg}(\theta)}(\boldsymbol{\epsilon}', T)$ \Comment{$\mathrm{sg}$ is stop-gradient operator}
\EndIf
\State $\mathrm{Sample}$ $\boldsymbol{\epsilon} \sim p(\boldsymbol{\epsilon})$, $t \sim p(t)$, $r \sim p(r \mid t, \mathrm{Iters})$
\State $\mathrm{Compute}$ $\x_t = \x_0 + t \cdot \boldsymbol{\epsilon}$, $\Delta t = t-r$
\State $\mathrm{Compute}$ $\x_r = \text{Solver}(\mathbf{x}_t, t \to r, f_\phi)$
\State $L(\theta) = w(t) \cdot \dist(f_\theta(\x_t), f_{\operatorname{sg}(\theta)}(\x_r))$
\State $\theta \leftarrow \theta - \eta \nabla_\theta L(\theta)$
\State $\mathrm{Iters} = \mathrm{Iters} + 1$
\Until{$\Delta t \to \mathrm{d}t$}
\Return $\theta$ \Comment{ECD}
\end{algorithmic}
\end{algorithm}

\section{Experimental Details} \label{sec:experiment-details}

\paragraph{Model Setup.} 

For both unconditional and class-conditional CIFAR-10 experiments, we initial ECMs from the pretrained EDM~\citep{Karras2022edm} of DDPM++ architecture~\citep{scoresde}.
For class-conditional \imgnet\ experiments, we initial ECM-{S/M/L/XL}, ranging from $280\mathrm{M}$ to $1.1\mathrm{B}$, from the pretrained EDM2~\citep{Karras2024edm2}. 
Detailed model configurations are presented in \cref{tab:experimental-config}. 

We follow \citep{Karras2022edm,cm} and set $c_{\text{skip}}(t) = \nicefrac{\sigma_{\text{data}}^2}{(t^2 + \sigma_{\text{data}}^2)}$ and $c_{\text{out}}(t) = \nicefrac{t \sigma_{\text{data}}}{\sqrt{t^2 + \sigma_{\text{data}}^2}}$, where $\sigma_{\text{data}}^2$ is the variance of (normalized) data, and set to $0.5$ for both CIFAR-10 and \imgnet.  

\paragraph{Computational Cost.} ECT is computationally efficient. On \imgnet, the tuning stage of ECT requiring only $0.39\%$ of the iCT~\citep{ict} training budget, and $0.60\%$ to $1.91\%$ of the EDM2~\citep{Karras2024edm2} pretraining budget depending on the model sizes. 
The exact computational resources required to train each individual model are shown in \cref{tab:experimental-config}.

\begin{table}[t]
\centering
\caption{Model Configurations and Training Details for unconditional and class-conditional ECMs on CIFAR-10, and ECM-S/M/L/XL on \imgnet.}
\vspace{3pt}
\begin{adjustbox}{max width=\linewidth}
\begin{tabular}{lcccccc}
\toprule
\multirow{2}{*}{\textbf{Model Setups}} & \multirow{2}{*}{Uncond CIFAR-10}  & \multirow{2}{*}{Cls-Cond CIFAR-10} & \multicolumn{4}{c}{\imgnet} \\
 \cmidrule(lr){4-7}
 &  &  & ECM-S & ECM-M & ECM-L & ECM-XL \\
\midrule
Model Channels  & $128$ & $128$ & $192$ & $256$ & $320$ & $384$ \\
Model capacity (Mparams) & $55.7$ & $55.7$ & $280.2$ & $497.8$ & $777.5$ & $1119.3$ \\
Model complexity (GFLOPs) & $21.3$ & $21.3$ & $101.9$ & $180.8$ & $282.2$ & $405.9$ \\
\midrule
\multicolumn{7}{l}{\textbf{Training Details}}\\
\midrule
Training Duration (Mimg) & $12.8$ & $12.8$ & $12.8$ & $12.8$ & $12.8$ & $12.8$ \\
Minibatch size  & $128$   & $128$  & $128$ & $128$   & $128$    & $128$ \\
Iterations  & $100$k & $100$k  & $100$k & $100$k   & $100$k    & $100$k \\
Dropout probability & $20\%$ & $20\%$ & $40\%$ & $50\%$ & $50\%$ & $50\%$   \\
Dropout feature resolution & - & - & $\leq 16\times16$ & $\leq 16\times16$ & $\leq 16\times16$ & $\leq 16\times16$ \\
Optimizer & RAdam & RAdam & Adam & Adam & Adam & Adam \\
Learning rate max ($\alpha_{\text{ref}}$) & $0.0001$ & $0.0001$ & $0.0010$ & $0.0009$ & $0.0008$ & $0.0007$ \\
Learning rate decay ($t_{\text{ref}}$) & - & - & $2000$ & $2000$ & $2000$ & $2000$ \\
EMA beta    & $0.9999$ & $0.9999$ & - & - & - & - \\
\midrule
\multicolumn{7}{l}{\textbf{Training Cost}}\\
\midrule
Number of GPUs  & $1$ & $1$ & $4$ & $8$ & $8$ & $8$ \\
GPU types       & A6000 & A6000 & H100 & H100 & H100 & H100 \\
Training time (hours) & $24$ & $24$ & $8.5$ & $8.5$ & $12$ & $15$ \\
\midrule
\multicolumn{7}{l}{\textbf{Generative Performance}} \\
\midrule
$1$-step FID         &  4.54  &  3.81  &  5.51  &  3.67  &  3.55  &  3.35   \\
$2$-step FID         &  2.20  &  2.02  &  3.18  &  2.35  &  2.14  &  1.96   \\
\midrule
\multicolumn{7}{l}{\textbf{ECT Details}} \\
\midrule
Regular Weighting ($\Bar{w}(t)$) & $\nicefrac{1}{(t-r)}$ & $\nicefrac{1}{(t-r)}$ & $\nicefrac{1}{t^2}+\nicefrac{1}{\sigma_{\text{data}}^2}$ & $\nicefrac{1}{t^2}+\nicefrac{1}{\sigma_{\text{data}}^2}$ & $\nicefrac{1}{t^2}+\nicefrac{1}{\sigma_{\text{data}}^2}$ & $\nicefrac{1}{t^2}+\nicefrac{1}{\sigma_{\text{data}}^2}$ \\
Adaptive Weighting ($w(\Delta)$) & \checkmark & \checkmark & \checkmark & \checkmark & \checkmark & \checkmark \\
Adaptive Weighting Smoothing ($c$) & $0.0$ & $0.0$ & $0.06$ & $0.06$ & $0.06$ & $0.06$ \\
Noise distribution mean ($P_{\text{mean}}$) & $-1.1$ & $-1.1$ & $-0.8$ & $-0.8$ & $-0.8$ & $-0.8$ \\
Noise distribution std ($P_{\text{std}}$) & $2.0$ & $2.0$ & $1.6$ & $1.6$ & $1.6$ & $1.6$ \\
\bottomrule
\end{tabular}
\end{adjustbox}
\label{tab:experimental-config}
\end{table}
\vspace{0.3cm}

\paragraph{Training Details.} 
We use RAdam~\citep{radam} optimizer for experiments on CIFAR-10 and Adam~\citep{adam} optimizer for experiments on \imgnet. 
We set the $\beta$ to ($0.9$, $0.999$) for CIFAR-10 and ($0.9$, $0.99$) for \imgnet. All the hyperparameters are indicated in \cref{tab:experimental-config}.
We do not use any learning rate decay, weight decay, or warmup on CIFAR-10. We follow EDM2~\citep{Karras2024edm2} to apply an inverse square root learning rate decay schedule on \imgnet.

On CIFAR-10, we employ the traditional Exponential Moving Average (EMA). To better understand the influence of the EMA rate, we track three Power function EMA~\citep{Karras2024edm2} models on \imgnet, using EMA lengths of $0.01$, $0.05$, and $0.10$. The multiple EMA models introduce no visible cost to the training speed. Considering our training budget is much smaller than the diffusion pretraining stage, we didn't perform Post-Hoc EMA search as in EDM2~\citep{Karras2024edm2}.

Experiments for ECT are organized in a non-adversarial setup to better focus and understand CMs and avoid inflated FID~\citep{fd_dino}.
We conducted ECT using full parameter tuning in this work, even for models over $1\mathrm{B}$ parameters. 
Investigating the potential of Parameter Efficient Fine Tuning (PEFT)~\citep{hu2021lora} can further reduce the cost of ECT to democratize efficient generative models, which is left for future research.

We train multiple ECMs with different choices of batch sizes and training iterations. 
By default, experiments on ImageNet 64$\times$64 utilize a batch size of $128$ and $100$k iterations, leading to a training budget of $12.8\mathrm{M}$.
We have individually indicated other training budgets alongside the relevant experiments, wherever applicable. 

\paragraph{Sampling Details.}
We apply stochastic sampling for 2-step generation.
For 2-step sampling, we follow \citep{ict} and set the intermediate $t=0.821$ for CIFAR-10, and $t=1.526$ for \imgnet. 

Intriguingly, these sampling schedules, originally developed for iCT, also perform well with our ECMs. This effectiveness across different CMs and training methods likely links to the inherent characteristics of the datasets and the forward process. Developing a scientific approach to determine optimal intermediate sampling schedules for CMs remains an open research problem.

\paragraph{Evaluation Metrics.} For both CIFAR-10 and \imgnet, FID and \dino\ are computed using $50$k images sampled from ECMs. As suggested by recent works~\citep{fd_dino,Karras2024edm2}, \dino\ aligns better with human evaluation. We use \texttt{dgm-eval}\footnote{\href{https://github.com/layer6ai-labs/dgm-eval/tree/master}{https://github.com/layer6ai-labs/dgm-eval}} to calculate \dino~\citep{fd_dino} to ensure align with previous practice.

\paragraph{Visualization Setups.} Image samples in \cref{fig:imgnet-scaling-samples} are from class \texttt{bubble} (971), class \texttt{flamingo} (130), class \texttt{golden retriever} (207), class \texttt{space shuttle} (812), classs \texttt{Siberian husky} (250), classs \texttt{ice cream} (928), class \texttt{oscilloscope} (688), class \texttt{llama} (355), class \texttt{tiger shark} (3).

Each triplet (left-to-right) includes from $2$-step samples from ECM-S trained with $12.8\mathrm{M}$ images, ECM-S trained with $102.4\mathrm{M}$ images, and ECM-XL trained with $102.4\mathrm{M}$ images.

\paragraph{1 GPU Hour Prototyping Settings.} This configuration uses a fixed $\Delta t \approx \mathrm{d}t$ by setting $q=256$ in our mapping function (corresponding to $\frac{r}{t} \approx 0.99$) and an EMA rate of 0.9993 for the model parameters. Using these settings, we run $8000$ gradient descent steps with a batch size of 128 on a single A100 40GB GPU. 

\paragraph{Scaling of Training Compute.}
For the results on scaling laws for training compute on CIFAR-10 shown in \cref{fig:imgnet-scaling} (Left), we train 6 class-conditional ECMs, each with varying batch size and number of training iterations. 
All ECMs in this experiment are initialized from the pretrained class-conditional EDM.

The minimal training compute at $2^0$ scale corresponds to a total budget of $12.8\mathrm{M}$ training images. The largest training compute at $2^5$ scale utilizes a total budget of $409.6\mathrm{M}$ training images, at $2\times$ EDM pretraining budget.

The first two points of $2^0$ and $2^1$ on \cref{fig:imgnet-scaling} (Left) use a batch size of $128$ for $100$k and $200$k iterations, respectively.
The third point of $2^2$ corresponds to ECM trained with batch sizes of $256$ for $200$k iterations. 
The final three points of $2^3$, $2^4$, and $2^5$ correspond to ECM trained with a batch size of $512$ for $200$k, $400$k, and $800$k iterations, respectively, with the smoothing factor $c=0.03$ enabled in the adaptive weighting $w(\Delta)$.
We use $\Bar{w}(t) = \nicefrac{1}{t}$ as the timestep weighting function to train all these models as this $\Bar{w}(t)$ achieves good performance on \dino.

\paragraph{Scaling of Model Size and Model FLOPs.}
We include details of model capacity as well as FLOPs in \cref{tab:experimental-config} to replicate this plot on \imgnet. 

On \imgnet, we scale up the training budgets of ECM-S and ECM-XL from $12.8\mathrm{M}$ (batch size of $128$ and $100$k iterations) to $102.4\mathrm{M}$ (batch size of $1024$ and $100$k iterations). We empirically find that scaling the base learning rate by $\sqrt{n}$ works well when scaling the batch size by a factor of $n$ when using Adam~\citep{adam} optimizer.

\section{Qualitative Results} \label{sec:visual-results}
\begin{figure}[h!]
    \vspace{5pt}
    \centering
    \includegraphics[width=\linewidth]{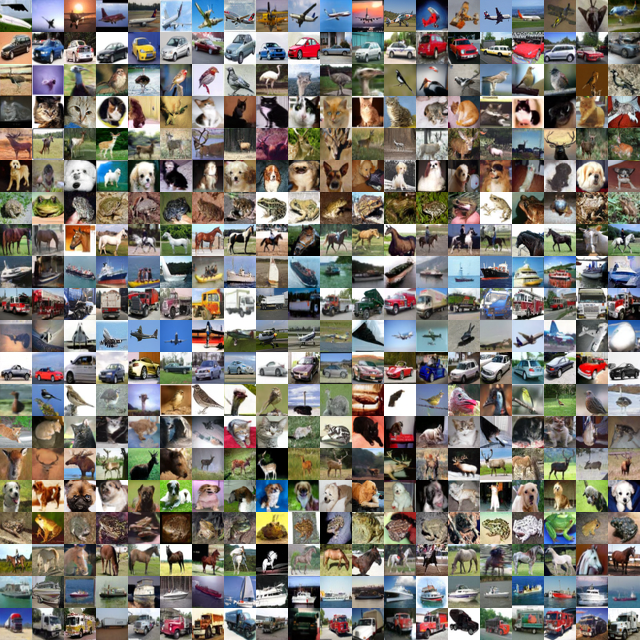}
    \caption{2-step samples from class-conditional ECM trained on CIFAR-10. Each row corresponds to a different class.}
    \label{fig:cifar-2-step}
\end{figure}

\begin{figure}[h!]
    \centering
    \includegraphics[width=\linewidth]{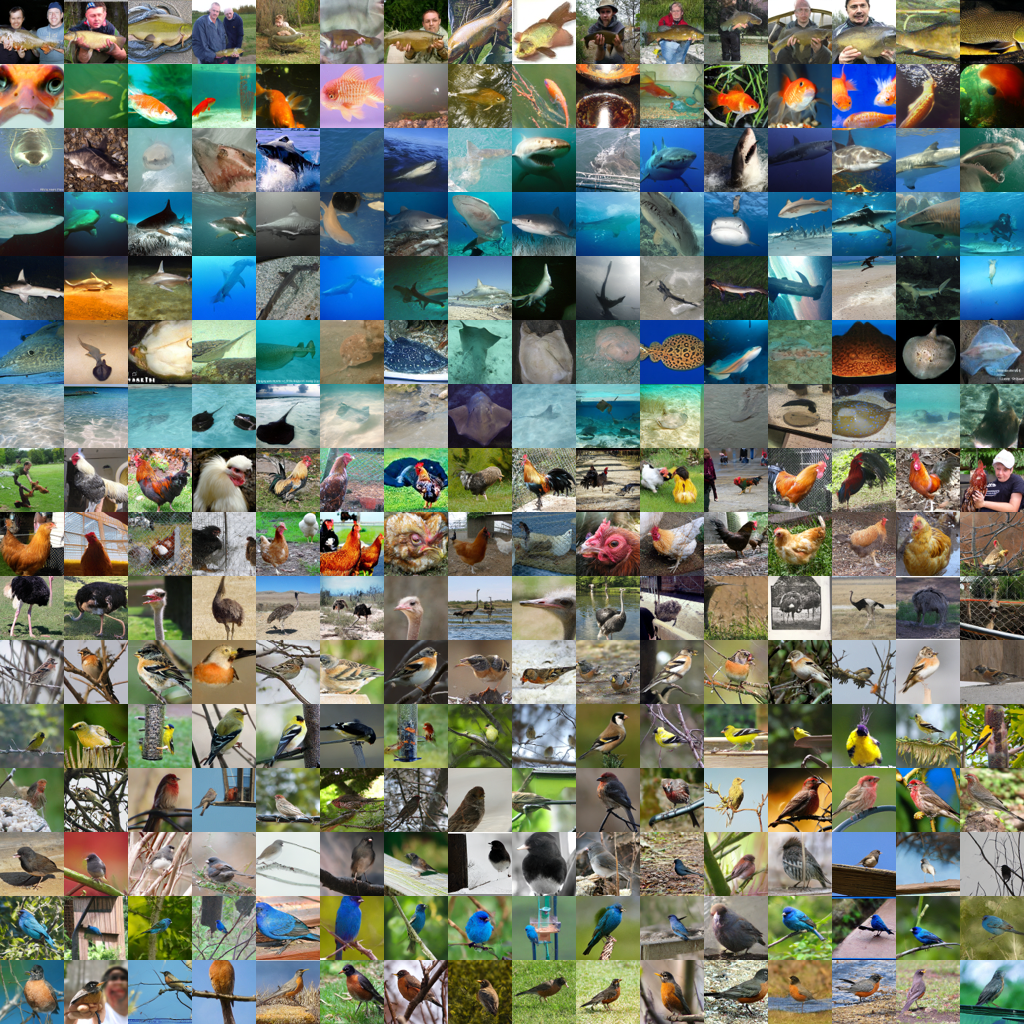}
    \caption{2-step samples from class-conditional ECM-XL trained on \imgnet. Each row corresponds to a different class.}
    \label{fig:imagenet-2-step}
\end{figure}

We provide some randomly generated 2-step samples from ECMs trained on CIFAR-10 and ImageNet-$64\times64$ in \cref{fig:cifar-2-step} and \cref{fig:imagenet-2-step}, respectively.

%%%%%%%%%%%%%%%%%%%%%%%%%%%%%%%%%%%%%%%%%%%%%%%%%%%%%%%%%%%%%%%%%%%%%%%%%%%%%%%
%%%%%%%%%%%%%%%%%%%%%%%%%%%%%%%%%%%%%%%%%%%%%%%%%%%%%%%%%%%%%%%%%%%%%%%%%%%%%%%
% APPENDIX
%%%%%%%%%%%%%%%%%%%%%%%%%%%%%%%%%%%%%%%%%%%%%%%%%%%%%%%%%%%%%%%%%%%%%%%%%%%%%%%
%%%%%%%%%%%%%%%%%%%%%%%%%%%%%%%%%%%%%%%%%%%%%%%%%%%%%%%%%%%%%%%%%%%%%%%%%%%%%%%

\end{document}